\documentclass[a4paper,fleqn]{cas-dc}

\usepackage[authoryear,longnamesfirst]{natbib}
\usepackage{amsmath}
\usepackage{amssymb}
\usepackage{booktabs}
\usepackage{multirow}
\usepackage{xcolor}
\usepackage{graphicx} 
\usepackage{microtype}
\newif\ifanonymous
\anonymousfalse   

\newcommand{\ours}{MPACT}

\newif\ifshowrevisions
\showrevisionstrue
\showrevisionsfalse
\newcommand{\rev}[1]{\ifshowrevisions\textcolor{blue}{#1}\else#1\fi}

\newcommand{\stk}[1]{\begin{tabular}[t]{@{}l@{}}#1\end{tabular}}
\newcommand{\sd}[1]{\,{\scriptsize$\pm$\,#1}}
\newcommand{\bvec}[1]{\boldsymbol{#1}}
\newlength{\qualcolw}
\newlength{\qualimgw}
\newlength{\qualimgh}
\newcommand{\qualimg}[1]{%
  \makebox[\qualcolw][c]{%
    \includegraphics[width=\qualimgw,height=\qualimgh,keepaspectratio]{#1}}}

\begin{document}
\let\WriteBookmarks\relax
\def\floatpagepagefraction{1}
\def\textpagefraction{.001}

\shorttitle{Source-Free Domain Adaptation for Multi-Label Classification}
\ifanonymous
\shortauthors{Anonymous}
\else
\shortauthors{S.N. B\"uy\"ukba\c{s} et~al.}
\fi

\title[mode=title]{Source-Free Domain Adaptation for Multi-Label Image
Classification: A Framework, Baselines, and Benchmark}

\ifanonymous
\author[1]{Anonymous}
\affiliation[1]{organization={Author details withheld for double-anonymized review}}
\else
\author[1]{Salih Numan B\"uy\"ukba\c{s}}[orcid=0009-0002-3347-3931]
\cormark[1]
\ead{salihnuman@sabanciuniv.edu}
\credit{Conceptualization, Methodology, Software, Investigation, Writing -- original draft}

\author[1]{Erchan Aptoula}[orcid=0000-0001-6168-2883]
\ead{erchan.aptoula@sabanciuniv.edu}
\credit{Conceptualization, Supervision, Writing -- review \& editing}

\author[2]{Mamatha Thota}[orcid=0000-0001-6553-4028]
\ead{MThota@lincoln.ac.uk}
\credit{Writing -- review \& editing}

\author[3]{Petra Bosilj}[orcid=0000-0001-9640-9828]
\ead{petra.bosilj@maastrichtuniversity.nl}
\credit{Writing -- review \& editing}

\affiliation[1]{organization={VPALab, Faculty of Engineering and Natural Sciences, Sabanci University},
            addressline={Orta Mahalle},
            city={Tuzla, Istanbul},
            postcode={34956},
            state={},
            country={T\"urkiye}}

\affiliation[2]{organization={Applied Vision and Artificial Intelligence Lab, University of Lincoln},
            addressline={Brayford Pool},
            city={Lincoln},
            postcode={LN6 7TS},
            state={Lincolnshire},
            country={United Kingdom}}

\affiliation[3]{organization={Department of Advanced Computing Sciences, Maastricht University},
            addressline={Paul-Henri Spaaklaan 1},
            city={Maastricht},
            postcode={6229 EN},
            state={Limburg},
            country={The Netherlands}}

\cortext[1]{Corresponding author}
\fi

\begin{abstract}
Source-free domain adaptation (SFDA) adapts a model trained on a labelled
source domain to an unlabelled target domain without access to the source
data. Almost the entire SFDA literature assumes mutually exclusive classes,
yet many deployed settings are multi-label. This paper formulates SFDA for
multi-label classification and establishes the first benchmark for it: five
adaptation tasks over six datasets spanning remote sensing, natural,
clipart-style and driving imagery, with four single-label SFDA methods
extended to provide the first baselines and a published multi-label method run
alongside them, over three seeds. The benchmark shows that extending a
single-label method is unreliable rather than merely suboptimal: half of the
extended baselines fail to improve on the unadapted source model at all, and
on two of the five tasks none of them does. The failures are not incidental.
Nearest-centroid pseudo-labelling, softmax-based confidence, the diversity
term of information maximization and confidence-weighted prototype
construction each rely on mutual exclusivity, and each fails differently
without it, so replacing the softmax with per-class sigmoids is a valid but
insufficient extension. Multi-label Prototype Alignment with Contrastive
Teaching (\ours{}) is proposed in response, a mean-teacher framework whose
objective, prototype construction and alignment term are each reconsidered for
overlapping label sets. It attains the best average on every reported metric
and is the only method that improves on the unadapted source model on all five
tasks. The analysis further introduces a prototype-agreement diagnostic
separating feature-space alignment from self-reinforcing agreement with the
classifier, and reports a negative result for confidence-based pseudo-label
filtering.
\end{abstract}

\begin{highlights}
\item \rev{First multi-label source-free domain adaptation benchmark for images}
\item Four single-label SFDA methods extended to multi-label as the first baselines
\item Benchmarked on five adaptation tasks across six multi-label datasets
\item \rev{Prototype contrastive alignment pays on two of five transfers and costs on one}
\item A prototype-agreement AUC diagnostic separates real alignment from feedback
\end{highlights}

\begin{keywords}
Source-free domain adaptation \sep Multi-label classification \sep Pseudo-labeling \sep
Self-training
\end{keywords}

\maketitle

\section{Introduction}\label{sec:intro}

\rev{The performance of deep classifiers drops when the distribution that
produced their test data differs from the distribution that produced their
training data}~\rev{\citep{torralba2011bias}}. Unsupervised domain adaptation (UDA) addresses this by
exploiting labeled source data and unlabeled target data~\rev{\citep{ganin2015unsupervised}}, but the
assumption that source data remains available at adaptation time is often
not correct in practice: medical and satellite imagery is often licensed for a
single use, commercial datasets cannot be redistributed with a deployed model,
and online deployments rarely have enough storage or bandwidth to carry a large training
corpus. Source-free domain adaptation (SFDA) removes that assumption, adapting
using only the pre-trained source model and unlabeled target data~\rev{\citep{liang2020shot}}.

SFDA research has developed rapidly, but \rev{almost} the entire literature assumes
\emph{single-label} classification~\rev{\citep{liang2020shot,acsfda}}. This assumption is deeply embedded rather
than incidental. SHOT-style methods obtain pseudo-labels by assigning each
target sample to \rev{whichever class centroid lies closest to it}~\citep{liang2020shot}; information-maximization regularizers encourage
a near-uniform marginal over a set of mutually exclusive classes;
confidence measures are computed from a softmax distribution whose entries (i.e predictions for each class)
compete by construction. However, many of the settings practitioners actually face are
multi-label: an aerial image contains \emph{buildings}, \emph{trees} and \emph{pavement}
simultaneously; a driving scene contains \emph{car}, \emph{person} and
\emph{traffic sign} at once. \rev{A single-label SFDA method can be applied to such data by swapping the
softmax for per-class sigmoids, and that substitution is a valid extension. It
is not, however, a sufficient one: a method ported this way may fail to improve
on the unadapted source model, and can perform worse than leaving it
alone.}

\rev{This paper argues that the extension is not trivial, because four
mechanisms that single-label SFDA relies on do not carry over unchanged once
classes may co-occur: softmax-based confidence, nearest-centroid
pseudo-labelling, the diversity target of information maximization, and the
treatment of label noise as a single mistaken class.
Section~\ref{sec:problem} names the properties of the setting on which each
depends, and Section~\ref{sec:method} takes them in turn.}

This paper makes the following contributions:

\begin{enumerate}
    \item \rev{Source-free domain adaptation is formulated for multi-label
          classification, making explicit which assumptions of the single-label
          formulation must be relaxed and why} (Section~\ref{sec:problem}).
    \item The first multi-label SFDA benchmark for images is introduced,
          covering five adaptation tasks over six datasets spanning remote
          sensing, natural and clipart-style images, and driving scenes. Four
          representative single-label SFDA methods are extended to provide the
          first baselines and are run alongside NOTELA~\citep{notela}, a
          published multi-label SFDA method, with every entry averaged over
          three seeds (Sections~\ref{sec:setup}--\ref{sec:results}).
    \item Multi-label Prototype Alignment with Contrastive Teaching
          (\ours{}) is proposed: a mean-teacher framework in which the
          objective, the prototype construction and the alignment term are
          each reconsidered for overlapping label sets, contributing a
          threshold-aware prototype initialization and a class-conditioned
          weighted alignment term. It attains the best average on every
          reported metric and is the only method to improve on the unadapted
          source model on all five tasks (Section~\ref{sec:method}).
    \item \rev{A prototype-agreement diagnostic is introduced that separates
          genuine feature--prototype alignment from self-reinforcing agreement
          with the classifier} (Section~\ref{sec:contrastive})\rev{, together
          with a negative result on confidence-based pseudo-label filtering}
          (Section~\ref{sec:negative}).
\end{enumerate}

\section{Related Work}\label{sec:related}

\paragraph{Multi-label domain adaptation.}
Multi-label UDA has received limited but growing attention. Approaches divide
into those that train a domain discriminator, such as DA-MAIC~\citep{damaic},
and discriminator-free ones such as DALN~\citep{daln} and
DDA-MLIC~\citep{ddamlic}.
DDA-MLIC derives its adversarial critic from the task classifier instead of an auxiliary subnet, with the reasoning that separating classification
from discrimination weakens both. Such
methods retain access to source data throughout adaptation and therefore serve
as a source-accessing reference rather than a comparable baseline, since the
setting studied here withholds source data entirely. This gap is reported
explicitly (Sections~\ref{sec:benchmark}
and~\ref{sec:discussion}).

\paragraph{Source-free domain adaptation.}
\rev{Deprived of source data, SFDA methods must substitute some other signal
for the supervision they have lost, and they differ in what that signal
is. One family derives pseudo-labels from the structure of the target feature
space while regularizing the output distribution: SHOT~\citep{liang2020shot}
freezes the source classifier, adapts only the feature extractor, and pairs
information maximization with pseudo-labels obtained from class centroids,
while AC-SFDA~\citep{acsfda} adds an adaptive curriculum over target samples. A
second family enforces agreement between each sample and its neighbours in
feature space~\citep{nrc}. NVC-LLN~\citep{nvclln}
belongs here, treating adaptation as learning with label noise. \rev{So does
UCON~\citep{ucon}: its contrastive term attracts each sample to its nearest
neighbours and repels it from the rest of the batch, with an
augmentation-consistency term and a partial-label loss added to control the
uncertainty in that supervision.} A third substitutes
self-supervision for the missing labels, typically a consistency or contrastive
objective over differently augmented views of the same target
image~\citep{csfda}. A fourth
maintains a teacher as a temporal ensemble of the student and trains on its
denoised predictions~\citep{cleanadapt}, the family to which the
framework proposed here belongs.
SHOT, UCON, NVC-LLN and AC-SFDA are the four methods extended and benchmarked in this paper.}

\paragraph{\rev{Multi-label source-free adaptation.}}
\rev{Two works address multi-label classification without source data.
NOTELA~\citep{notela} augments a denoising teacher--student scheme with a
Laplacian term encouraging neighbouring points in feature space to receive
similar pseudo-labels, and evaluates it on multi-label bioacoustic recordings.
It reports that existing SFDA methods there can perform worse than no
adaptation at all, and that restricting the data to single-label recordings
recovers much of their performance -- independent evidence, in another
modality, for the premise of this paper. Its own treatment of the multi-label
case is nonetheless the generic one, each class handled as an independent
binary problem, and its image benchmarks are single-label.
CoWA~\citep{cowa} adapts a multi-label chest X-ray classifier at test time,
weighting each sample by how far its predictions depart from the label
co-occurrence structure estimated from the target data; it updates only the
normalization parameters, in a single streaming pass. It addresses
test-time adaptation, in which the target data arrive as a stream and the model
is revised as they pass; the problem posed here is source-free domain
adaptation, in which the unlabelled target set is available in full and
offline. The two are neighbouring formulations but not the same, and a
method written for either carries assumptions the other does not make. The
setting studied here
differs from both: multi-label image classification, with the full unlabelled
target set available offline and the entire representation adapted over
multiple epochs.}

\paragraph{\rev{Prototypes and teacher--student adaptation.}}
\rev{Two components of the framework proposed here are taken from the
single-label literature, not invented here: class prototypes and
a teacher--student pair. Class prototypes are widely used to structure a
target feature space: as
centroids for pseudo-labelling~\citep{liang2020shot}, as targets for
distribution alignment~\citep{cpd}, and as carriers of class information added
directly to target features to compensate for the domain shift~\citep{pfc}. The Class Prototype Discovery construction of~\citet{cpd} is
the one adopted here.
The teacher--student scheme itself
originates outside domain adaptation, as a semi-supervised
method~\citep{tarvainen2017mean}; CleanAdapt-TS~\citep{cleanadapt} shows for
source-free video action recognition that such a teacher yields more consistent
pseudo-labels than the source model alone, and AC-SFDA~\citep{acsfda} and
NOTELA~\citep{notela} maintain one as well.}

\paragraph{Learning with noisy labels.}
Because target pseudo-labels are noisy by construction, losses that bound the
influence of a mislabelled example are directly relevant. \rev{The standard
approaches replace or temper the cross-entropy term: the mean absolute error is
robust to label noise but slow to converge, and generalized cross entropy
interpolates between the two~\citep{mae,generalizedce};
others reweight suspect samples rather than changing the
loss~\citep{reweighting}.} Symmetric cross
entropy~\citep{wang2019sce} instead combines a
forward and a reverse term to bound the influence of mislabeled samples. Its binary, per-class analogue is used in the proposed framework (described in Section~\ref{sec:sbce}).

\section{Problem Formulation}\label{sec:problem}

Let $\mathcal{X}$ denote the input space and $C$ the number of classes. In
multi-label classification an image $\bvec{x}\in\mathcal{X}$ carries a binary label
vector $\bvec{y}\in\{0,1\}^{C}$\rev{, whose entries $y_{1},\dots,y_{C}$ take
$y_{c}=1$ when class $c$ is present}. Any subset of the $C$ classes may occur
together, so $0 \leq \lVert\bvec{y}\rVert_{1} \leq C$\rev{, in contrast to the
single-label problem, in which $\bvec{y}$ is restricted to the $C$ one-hot
vectors}.

A source model $\Theta_{s}=g_{\phi}\circ h_{\psi}$ is trained on a
source dataset $D_{s}=\{(\bvec{x}^{s}_{i},\bvec{y}^{s}_{i})\}_{i=1}^{n_{s}}$
\rev{of $n_{s}$ labelled samples,} drawn from a source distribution $P_{s}$,
where the encoder
$h_{\psi}:\mathcal{X}\to\mathbb{R}^{D}$ produces a feature $\bvec{z}=h_{\psi}(\bvec{x})$,
\rev{with $D$ the dimensionality of the representation fixed by the backbone,}
and the classifier head
$g_{\phi}:\mathbb{R}^{D}\to\mathbb{R}^{C}$ produces per-class logits, converted
to independent probabilities \rev{$\bvec{q}=\sigma(g_{\phi}(\bvec{z}))\in[0,1]^{C}$}
by an elementwise sigmoid $\sigma$. A target
dataset $D_{t}=\{\bvec{x}^{t}_{j}\}_{j=1}^{n_{t}}$ \rev{of $n_{t}$ samples} is
drawn from a distribution $P_{t}\neq P_{s}$ over the same $C$ classes.
\rev{Ground-truth labels $\bvec{y}^{t}_{j}$ exist for these samples, but they
are withheld throughout adaptation and are used only to evaluate the adapted
model.}

The source-free constraint states that adaptation has access to $\Theta_{s}$ and
$D_{t}$ only. The source data are unavailable, \rev{as is any
external summary of them kept for transfer, such as stored source features or
class statistics. The source model's own parameters and buffers are
inherited, as in all source-free adaptation; this includes
batch-normalization running statistics, which start from their source values
and are updated on target batches during adaptation.} The task is
to produce an adapted model $\Theta_{t}$ that performs well on $P_{t}$.

\rev{Three properties of this setting drive the design that follows:}

\rev{\textbf{(P1) Non-exclusivity.} Any subset of the $C$ classes may be
present, so the model's per-class outputs are independent Bernoulli
probabilities rather than shares of one distribution, without the
mutual-exclusivity constraint a softmax imposes.
Every construction by which single-label adaptation expresses competition among
classes -- a softmax over the logits, a one-of-$C$ cluster assignment, a
partition of a batch into one positive and $C-1$ negatives -- takes its
meaning from that coupling. Each therefore ceases to be defined under
multi-label semantics rather than only becoming less accurate.}

\rev{\textbf{(P2) Sparsity and imbalance.} Positives are rare and unevenly
distributed: across the datasets of Table~\ref{tbl:datasets} between $55\%$ and
$92\%$ of all (sample, class) entries are negative, and a handful of classes
carry most of the positives. Any per-class quantity estimated from a
mini-batch -- a marginal, a centroid, a confidence weight -- is therefore
computed from a large majority of negatives and a few positives.}

\rev{\textbf{(P3) No validation signal.} As $\bvec{y}^{t}$ is withheld, no
quantity that depends on it is available during adaptation, a validation score
among them. \rev{Supervised model selection is therefore not applicable.
Unsupervised criteria are not ruled out in principle, but the convention in
this literature is to report the final model after a fixed
budget~\citep{liang2020shot}, which is what this study does.} \rev{The
constraint shapes} the experimental protocol as well as the method
(Section~\ref{sec:design}).}

\rev{The first two determine what the method can do; the third determines what
the experiments can claim.}

\section{Method}\label{sec:method}

\begin{figure*}[t]
  \centering
  \includegraphics[width=0.9\textwidth]{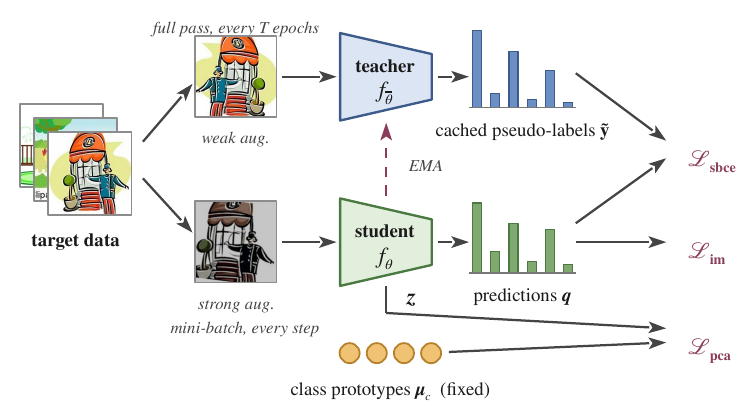}
  \caption{Overview of \ours{}. \rev{Every $T$ epochs the teacher
  $f_{\bar{\theta}}$ makes a full pass over the target data under weak
  augmentation and caches per-class soft pseudo-labels $\tilde{y}$; the
  student $f_{\theta}$ trains on mini-batches under strong augmentation and is
  supervised by that cache until the next refresh, through the symmetric term
  $\mathcal{L}_{\text{sbce}}$. The multi-label information maximization term
  $\mathcal{L}_{\text{im}}$ acts on the student's predictions $q$ alone. The student's
  pooled features $z$ are aligned to the class prototypes $\mu_{c}$ by
  $\mathcal{L}_{\text{pca}}$, with the prototypes constructed once before
  adaptation and held fixed throughout. The dashed arrow is the exponential
  moving average update of the teacher from the student, along which no
  gradient flows.}}\label{fig:pipeline}
\end{figure*}

\ours{} follows a mean-teacher self-training scheme\rev{, in the manner of
CleanAdapt-TS~\citep{cleanadapt}. What differs is the supervision that scheme
carries: a video action label is a single mutually exclusive class, whereas
each target image here contributes $C$ independent per-class targets of
differing reliability}. A student $f_{\theta}$ is
initialized from the source model and optimized on the target data; a teacher
$f_{\bar{\theta}}$ is maintained as an exponential moving average (EMA) of
the student,
$\bar{\theta}\leftarrow\alpha\,\bar{\theta}+(1-\alpha)\theta$ \rev{with decay
$\alpha$. Here $\theta=(\psi,\phi)$ collects the encoder and head parameters of
Section~\ref{sec:problem}, and $h_{\psi}$ below denotes the student's encoder
at its current value rather than the source one}. Every
$T$ epochs the teacher performs a full pass over the target set and writes
per-class soft targets $\tilde{y}_{jc}\in[0,1]$ \rev{for each target sample $j$
and class $c$}, which supervise the student
until the next refresh. The student is trained on strongly augmented views
while the teacher reads weakly augmented ones.
Figure~\ref{fig:pipeline} summarizes the pipeline.

Each of the three loss terms below, together with the prototype construction
that the second of them relies on, is obtained by taking the corresponding
single-label procedure, identifying the specific assumption it places on the
label space, and replacing that assumption rather than simply exchanging the
output nonlinearity. Every subsection follows that order: the standard
single-label form, the property of Section~\ref{sec:problem} whose withdrawal
breaks it, and the resulting formulation.

\subsection{Symmetric binary cross entropy}\label{sec:sbce}

\paragraph{Standard form.} Symmetric cross entropy~\citep{wang2019sce} pairs
the usual forward term with a reverse term in which the roles of prediction and
label are exchanged,
\begin{equation}
\mathcal{L}^{\text{sl}}_{\text{ce}} = -\sum_{k} y_{k}\log p_{k} ,
\qquad
\mathcal{L}^{\text{sl}}_{\text{rce}} = -\sum_{k} p_{k}\log y_{k} ,
\end{equation}
\rev{over $K$ mutually exclusive classes, with $p_{k}$ the predicted
probability of class $k$ and $y_{k}$ the one-hot label.} Because the one-hot
target is clamped away from zero, the reverse term is bounded, and it therefore limits how much a single
confidently mislabeled sample can contribute to the gradient. This is the
property that makes it attractive for self-training, where the targets are the
model's own noisy pseudo-labels.

\rev{\paragraph{What breaks.} The first failure is (P1). Under a softmax the
$K$ classes are covered by one
categorical distribution, and the reverse term above accounts for all of the
probability mass. Under independent sigmoids each class is instead its own
Bernoulli variable, carrying a positive and a negative outcome, so the
symmetric objective must be written per class.}

\rev{A second property fails less visibly. The bound quoted above assumes a
one-hot target, under which the reverse term collapses to a mean absolute
error and inherits that estimator's noise robustness~\citep{wang2019sce}. The
targets here are the teacher's soft sigmoid outputs, so no such collapse
occurs and the published justification does not transfer.}

\paragraph{Formulation.} \rev{With $q_{jc}$ the student's predicted
probability for sample $j$ and class $c$, and $\tilde{y}_{jc}$ the
corresponding teacher-derived target, both terms are averaged over the $B$
samples of the mini-batch and the $C$ classes,}
\begin{align}
\mathcal{L}_{\text{bce}} &= -\frac{1}{BC}\sum_{j=1}^{B}\sum_{c=1}^{C}
   \big[\tilde{y}_{jc}\log q_{jc} \notag \\
 &\qquad\qquad + (1-\tilde{y}_{jc})\log(1-q_{jc})\big], \\
\mathcal{L}_{\text{rbce}} &= -\frac{1}{BC}\sum_{j=1}^{B}\sum_{c=1}^{C}
   \big[q_{jc}\log \tilde{y}_{jc} \notag \\
 &\qquad\qquad + (1-q_{jc})\log(1-\tilde{y}_{jc})\big], \\
\mathcal{L}_{\text{sbce}} &= \mathcal{L}_{\text{bce}}
   + \mathcal{L}_{\text{rbce}} .
\end{align}
What bounds the reverse term here is instead the clamping of
$\tilde{y}_{jc}$ away from $0$ and
$1$. The term is therefore adopted for the behaviour it exhibits rather than by
appeal to the original analysis.

\rev{The binary form itself is not new.} GTNet~\citep{gtnet} introduces
symmetric binary cross entropy for human--object interaction detection, itself
a multi-label task, in order to absorb the confusing, mislabeled
annotations common in that setting. The present work adopts \rev{that objective
unchanged, both terms entering at equal weight as they do there}, but for a
different source of noise, which is the EMA teacher's predictions under domain shift.

\subsection{Prototype initialization}\label{sec:protoinit}

\paragraph{Standard form.} \rev{Prototype-based alignment anchors}
each class by a single prototype in feature space. Those prototypes are
constructed before adaptation begins by the Class Prototype Discovery
(CPD) procedure of~\citet{cpd}, which accumulates target features into one
prototype per class in a confidence-aware manner. For a mini-batch of
$B$ samples, CPD assigns sample $i$ a weight with respect to class $k$ and
updates the prototype by an exponential moving average \rev{with decay
$\eta$},
\begin{equation}
w_{i,k} = \frac{\exp(p_{i,k})}{\sum_{b=1}^{B}\exp(p_{b,k})},
\quad
\bvec{\mu}_{k} \leftarrow \eta\,\bvec{\mu}_{k}
  + (1-\eta)\sum_{i=1}^{B} w_{i,k}\,\bvec{z}_{i} ,
\label{eq:cpd}
\end{equation}
where $p_{i,k}$ is the score that sample $i$ belongs to class $k$, $\bvec{z}_{i}$
its feature and $\bvec{\mu}_{k}$ the prototype of class $k$. The design is deliberate on one point in particular: because the
weights are normalized across the batch rather than thresholded, every sample
contributes to every prototype, which \citet{cpd} contrast explicitly with
approaches ``limited to only high-confidence samples''.

\paragraph{What breaks.} Equation~\eqref{eq:cpd} rests on \rev{three}
properties of the
single-label setting. The first is stated in the original: the score vector is
a distribution over classes, $\sum_{k}p_{i,k}=1$. Under independent sigmoids
that identity does not hold, so the quantity being normalized is no longer a
share of a fixed unit of mass.

\rev{The second is implicit: that the prototype will be used by comparing it
against the others.} Since
$p_{i,k}\in[0,1]$, the exponential $\exp(p_{i,k})$ is confined to $[1,e]$, so the
most confident sample in a batch can outweigh the least confident by a factor
of at most $e\approx 2.72$. \rev{The weighting is therefore nearly uniform, and
Equation~\eqref{eq:cpd} returns approximately the mean feature of the batch,
displaced towards the samples carrying the class by an amount set by how many
of them there are. For a class occupying a small fraction of the batch that
displacement is small, and for a batch containing none of the class -- a probable case in general -- there is none at
all.}

\rev{The cap is not peculiar to sigmoids -- softmax scores lie in $[0,1]$
too -- but under mutual exclusivity it does little harm. The batch-mean part
of the prototype is shared by every class, and a sample is assigned by asking
which prototype is nearest, so what decides the outcome is how the prototypes
differ, and the shared part contributes little to that difference. No such
comparison is available here, since a sample may carry several classes, so the
shared part is never differenced away and features are drawn towards the batch
mean as much as towards the class. The weakness is therefore the same in both
settings; what (P1) removes is the comparison that kept it harmless. By (P2) the
displacements also differ in size from class to class, so the rarest classes,
which most need a well-placed anchor, receive the weakest one.}

\rev{A third property is implicit in the design rather than the arithmetic.
\citet{cpd} decline to threshold because, when every sample carries exactly one
label, a sample scoring low for class $k$ is still a positive for some other
class, so discarding it would discard information the batch statistics could
otherwise use. That premise is (P1). Under independent sigmoids a low score for
class $c$ is not weak evidence about where class $c$ lies in feature space; it
is evidence that the class is absent from that image.}

\paragraph{Formulation.} Two changes restore the intended behaviour. Scores are
read as independent per-class sigmoids, and the exponential normalization is
replaced by a linear one taken over the samples that exceed a confidence
threshold $\tau$, \rev{taken to be the sigmoid decision boundary},
\begin{equation}
\begin{split}
w_{i,c} &=
\frac{p_{i,c}\,\mathbf{1}[p_{i,c}>\tau]}
     {\sum_{b=1}^{B} p_{b,c}\,\mathbf{1}[p_{b,c}>\tau]} , \\[6pt]
\bvec{\mu}_{c} &\leftarrow \eta\,\bvec{\mu}_{c}
  + (1-\eta)\sum_{i=1}^{B} w_{i,c}\,\bvec{z}_{i} ,
\end{split}
\end{equation}
\rev{A} prototype whose class has no sample above
threshold in a batch is left unchanged\rev{, which is the empty case above:
with no positive to weight, any update is an update towards images lacking the
class. Prototypes are initialized to zero, so a class with no confident sample
anywhere in the target set keeps a zero prototype. Cosine similarity at the
origin is taken to be zero, so such a class contributes a constant term to
Equation~\eqref{eq:pca} and no gradient to the features}. Each change answers one of the
failures above. The threshold restricts the sum to the samples the model
believes carry the class, so what contamination survives comes from false
positives rather than from every image lacking the class. \rev{The linear weighting then makes a survivor's influence
proportional to its score rather than to $\exp$ of it, so a sample at $0.9$
outweighs one at $0.5$ by $1.8$ where the exponential form, whose floor
$\exp(0)=1$ dilutes every difference, allows only $1.49$.}

\rev{Thresholding is what \citet{cpd} set out to avoid, and the departure is
deliberate. The second failure makes it necessary and the third makes it
admissible.} Thresholding here is the
consequence of positives and negatives being separately identifiable per class. \rev{That separability also fixes the value of $\tau$, since the sigmoid
boundary is where the model stops asserting the class.}

\subsection{Prototype contrastive alignment}\label{sec:contrastive-loss}

\paragraph{Standard form.} Prototype contrastive objectives anchor each class
by a representative feature vector and pull samples of that class towards it
while pushing all other samples away, typically through an InfoNCE term in
which a sample is a positive for exactly one prototype and a negative for the
remaining $K-1$,
\begin{equation}
\mathcal{L}^{\text{sl}}_{\text{proto}}
 = -\frac{1}{B}\sum_{i=1}^{B}
   \log\frac{\exp(\mathrm{sim}(\bvec{z}_{i},\bvec{\mu}_{k_{i}})/\kappa)}
            {\sum_{k=1}^{K}\exp(\mathrm{sim}(\bvec{z}_{i},\bvec{\mu}_{k})/\kappa)} ,
\label{eq:infonce}
\end{equation}
\rev{where $k_{i}$ is the single class carried by sample $i$, $\mathrm{sim}$ a
similarity between feature vectors, $\kappa$ a temperature, and the
normalization runs over the $K$ prototypes. The form belongs to the contrastive
representation-learning literature~\citep{pcl}: the prototype-based SFDA
methods above instead use prototypes for assignment~\citep{liang2020shot} or
distribution alignment~\citep{cpd}.}

\paragraph{What breaks.} \rev{This is (P1) again.} That construction assumes
the positives and negatives
of prototype $c$ form a partition of the batch, which mutual exclusivity
guarantees and multi-label semantics do not. An image containing both
\emph{car} and \emph{person} is a legitimate positive for two prototypes at
once, and assigning it as a negative for either injects a repulsion that the
labels contradict. Hard assignment is not only suboptimal here; it is
undefined.

\paragraph{Formulation.} One prototype $\bvec{\mu}_{c}\in\mathbb{R}^{D}$ per class is
initialized before adaptation from the source model's features on the target
set, by the procedure of the preceding subsection, and held \emph{fixed}
thereafter. \rev{Two things then change.} Hard assignment is replaced by a soft
attraction weighted by the agreement between the prototype's class indicator
$\bvec{e}_{c}$ \rev{-- the one-hot vector carrying a single one in position
$c$ --} and the sample's multi-hot target $\tilde{\bvec{y}}_{j}$\rev{; and the
normalization is taken over the samples for each prototype rather than over
the prototypes for each sample}\rev{,}
\begin{equation}
\mathcal{L}_{\text{pca}}
 = -\frac{1}{C}\sum_{c=1}^{C}
   \frac{\sum_{j=1}^{B} w_{cj}\log \pi_{cj}}{\sum_{j=1}^{B} w_{cj}} ,
\label{eq:pca}
\end{equation}
that is, a weighted average of log-probabilities under the prototype-anchored
distribution $\pi_{cj}$, with the label-agreement weight $w_{cj}$,
\begin{align}
\pi_{cj} &= \frac{\exp(\mathrm{sim}(\bvec{\mu}_{c},\bvec{z}_{j})/\kappa)}
                 {\sum_{i=1}^{B}\exp(\mathrm{sim}(\bvec{\mu}_{c},\bvec{z}_{i})/\kappa)} ,\\[2pt]
w_{cj}   &= \cos(\bvec{e}_{c},\tilde{\bvec{y}}_{j})
          = \frac{\tilde{y}_{jc}}{\lVert\tilde{\bvec{y}}_{j}\rVert_{2}} .
\end{align}
Here $\bvec{z}_{j}=h_{\psi}(\bvec{x}_{j})$ is the student feature of the strongly augmented
view\rev{, $\mathrm{sim}$ is cosine similarity, and every sum runs over the $B$
samples of the mini-batch}.

The label-agreement weight follows \citet{buyukbas2025cosine}, which
introduces the cosine similarity between multi-hot label vectors as the weight
of a supervised contrastive objective for multi-label remote sensing scene
classification. Two things differ in the present use. That formulation
contrasts pairs of samples under ground-truth annotation, whereas the objective
above is anchored on class prototypes, and $\tilde{\bvec{y}}_{j}$ here is the teacher's vector of
soft per-class probabilities. The
weight is therefore graded by the teacher's confidence rather than by hard label
membership, so no entry of $w_{cj}$ is ever exactly zero and every sample
participates, however weakly, in every prototype's objective.

The formulation has two consequences that the hard-assignment form cannot
express. First, each prototype is normalized over the batch rather than each
sample over the prototypes \rev{as in Equation~\eqref{eq:infonce}}, so the $C$
memberships of a single image are
entries in $C$ separate distributions rather than shares of one. A sample is
therefore a weighted positive for every prototype whose class it carries, and
raising its proximity to $\bvec{\mu}_{\text{car}}$ does not mechanically lower its
proximity to $\bvec{\mu}_{\text{person}}$; co-occurring labels are jointly satisfiable
instead of competing for a fixed unit of probability mass.
 Second, the normalization by $\lVert\tilde{\bvec{y}}_{j}\rVert_{2}$ makes a
label-dense image contribute less to each individual prototype than a
label-sparse one, which prevents the handful of images carrying many labels
from dominating every prototype's objective simultaneously. \rev{When the targets
happen to be one-hot, $w_{cj}$ collapses to the indicator, which recovers the
hard-weighted form of this objective but not
Equation~\eqref{eq:infonce}: the two normalize along different axes, so no
choice of weight turns one into the other. Equation~\eqref{eq:pca} is
therefore a class-conditioned sample-discrimination objective in the
prototype-contrastive family rather than a soft-label reading of the standard
form, and that normalization is what allows overlapping label sets in the
first place. Section~\ref{sec:ablations} measures the two against each other.}

\subsection{Multi-label information maximization}

\paragraph{Standard form.} Information maximization, as used by
SHOT~\citep{liang2020shot}, combines entropy minimization on individual
predictions -- which sharpens them -- with a diversity term equal to the
entropy of the batch-averaged softmax marginal, which is maximized when that
marginal is uniform over the $K$ classes,
\begin{equation}
\begin{split}
\mathcal{L}^{\text{sl}}_{\text{im}}
 &= \frac{1}{B}\sum_{i}\mathcal{H}(\bvec{s}_{i})
 - \mathcal{H}\!\left(\frac{1}{B}\sum_{i}\bvec{s}_{i}\right) , \\
\bvec{s}_{i} &= \mathrm{softmax}\!\left(g_{\phi}(\bvec{z}_{i})\right) ,
\end{split}
\label{eq:imsl}
\end{equation}
\rev{with $\mathcal{H}$ the entropy of a $K$-way categorical distribution.}
Under mutual exclusivity the
diversity term is a mild and well-motivated anti-collapse prior: it prevents
the degenerate solution in which every sample is assigned the same class.

\paragraph{What breaks.} \rev{Here the binding property is (P2).} With
independent sigmoids the marginal of class $c$ is
a Bernoulli variable, and its binary entropy is maximized at
$\bar{q}_{c}=0.5$. The diversity term therefore pushes the mean predicted
probability of \emph{every} class towards $0.5$, \rev{which for confident
predictions amounts to asserting the class in half of the images}. For a class whose true prevalence is a
few percent, this is no longer
a weak prior against collapse but a large and systematically wrong target. Of
the three objectives treated here, this is the one whose single-label
assumption transfers least well.

\paragraph{Formulation.} The objective is ported to independent classes in its
minimal faithful form, \rev{Equation~\eqref{eq:imsl} with both terms computed
per class},
\begin{equation}
\begin{split}
\mathcal{L}_{\text{im}} &=
   \underbrace{\frac{1}{BC}\sum_{j=1}^{B}\sum_{c=1}^{C}\mathcal{H}(q_{jc})}_{\text{confidence}}
 - \underbrace{\frac{1}{C}\sum_{c=1}^{C}\mathcal{H}(\bar{q}_{c})}_{\text{diversity}} , \\
\bar{q}_{c} &= \frac{1}{B}\sum_{j=1}^{B}q_{jc} ,
\end{split}
\end{equation}
where \rev{$q_{jc}$ is the student's predicted probability for sample $j$ and
class $c$, the sums run over the $B$ samples of the mini-batch and the $C$
classes, and} $\mathcal{H}$ \rev{is now the binary entropy of a single Bernoulli
marginal rather than the entropy of a $K$-way categorical}. \rev{The term is
retained in this form}, but the mismatch
identified above is not resolved by it.

\subsection{Training objective and design choices}\label{sec:design}

The total objective in the default configuration is
\begin{equation}
\mathcal{L} = \lambda_{\text{sbce}}\mathcal{L}_{\text{sbce}}
 + \lambda_{\text{pca}}\mathcal{L}_{\text{pca}}
 + \lambda_{\text{im}}\mathcal{L}_{\text{im}} .
\end{equation}
\rev{Only the encoder $h_{\psi}$ is adapted; the classifier head $g_{\phi}$ is
frozen throughout, so the gradient step and the EMA both act on $\psi$ alone.
Freezing the head preserves the source decision geometry, so any change in
target performance is attributable to the representation
alone~\citep{liang2020shot}.}

A fourth term $\mathcal{L}_{\text{pce}}$, which asks the frozen classifier to
assign each prototype to its own class, is defined in the implementation but is
inactive whenever the prototypes are static, and therefore contributes nothing
to the results reported here; it applies only to the learnable-prototype
variant examined in Section~\ref{sec:ablations}.

\section{Experimental Setup}\label{sec:setup}
This section summarizes the datasets and adaptation tasks used for the experiments, the implementation details for each extended baseline as well as the proposed one, the changes by which the single-label baselines are extended to multi-label, the evaluation metrics\rev{, and the experiments those metrics are reported for}.

\subsection{Datasets and adaptation tasks}

Evaluation covers five adaptation tasks from three application families, so
that conclusions are not specific to one type of domain gap.
Table~\ref{tbl:datasets} summarizes the datasets.

\emph{Remote sensing.} UCM~\citep{ucm} and AID~\citep{aid} are aerial scene
datasets, \rev{re-annotated with a shared vocabulary of 17 multi-labels
by~\citet{ucmmulti} and~\citet{aidmulti} respectively}. The two directions
UCM$\rightarrow$AID and AID$\rightarrow$UCM differ in sensor, resolution and
scene composition.

\emph{Natural to clipart imagery.} PASCAL VOC~\citep{voc} and
Clipart1k~\citep{inoue2018clipart} share 20 object categories but differ
drastically in low-level statistics and rendering style, giving
VOC$\rightarrow$Clipart and Clipart$\rightarrow$VOC.

\emph{Driving scenes.} Cityscapes~\citep{cordts2016cityscapes} and its
synthetically fogged counterpart Foggy Cityscapes~\citep{sakaridis2018foggy}
are converted to an 11-class multi-label problem following the protocol of
DDA-MLIC~\citep{ddamlic}, which excludes categories present in nearly every
image. This task isolates a pure appearance shift: the underlying scenes are
identical and only the imaging conditions change. \rev{Only the
Cityscapes$\rightarrow$Foggy direction is evaluated, following
DDA-MLIC~\citep{ddamlic}.}

\rev{The split files, the scripts that derive image-level labels from each
dataset's own annotations, and the evaluation code are released with the
implementation, so the benchmark can be reproduced rather than only
described.}

\begin{table}[t]
\caption{Multi-label datasets used in this study. Density is the mean number of
positive labels per image on the training split. \rev{Cityscapes and Foggy
Cityscapes share one row because the latter re-renders the same scenes under
synthetic fog and inherits their annotations
unchanged.}}\label{tbl:datasets}
\begin{tabular*}{\tblwidth}{@{}LCCCC@{}}
\toprule
Dataset & Classes & Train & Val & Density \\
\midrule
UCM              & 17 & 1680 & 420  & 3.30 \\
AID              & 17 & 2400 & 600  & 5.15 \\
VOC              & 20 & 2111 & 2221 & 1.56 \\
Clipart1k        & 20 & 500  & 500  & 1.77 \\
\rev{Cityscapes / Foggy} & 11 & 2975 & 500  & 5.01 \\
\bottomrule
\end{tabular*}
\end{table}

\subsection{Implementation details}

\rev{Throughout this study the reported metric is the \emph{student}'s.} All models use an ImageNet-pre-trained ResNet backbone~\citep{he2016resnet}
with the fully connected layer replaced by a $C$-way linear classifier.
\rev{The pooled backbone output is used as the feature $\bvec{z}$ without an
intervening bottleneck, so $D=2048$.}
ResNet-101 is reported as the primary configuration, matching the backbone used
by the multi-label UDA reference~\citep{ddamlic} so that the comparison stays consistent.
\ours{} is trained for 45 epochs with SGD at learning rate $5\times10^{-4}$,
batch size 64, cosine decay, EMA decay $0.995$, dropout $0.2$, and loss weights
$\lambda_{\text{sbce}}=0.5$, $\lambda_{\text{pca}}=0.5$,
$\lambda_{\text{im}}=0.5$, $\lambda_{\text{pce}}=0.1$\rev{, the last of these
effective only in the learnable-prototype variant of
Section~\ref{sec:ablations}, since the term it weights is inactive while the
prototypes are static}. \rev{The prototype
confidence threshold is $\tau=0.5$, the prototype EMA decay $\eta=0.8$, and the
temperature of the contrastive term $\kappa=0.1$.} Pseudo-labels are refreshed
every three epochs.

\rev{The source model is trained on the labelled source split for 50 epochs
with SGD at learning rate $10^{-2}$, momentum $0.9$, weight decay $10^{-4}$ and
batch size 32, under binary cross entropy; the same source checkpoint
initializes every method on a given task. Adaptation runs on the target
\emph{training} split without its labels and is evaluated on the target
validation split, whose sizes Table~\ref{tbl:datasets} gives. The weak view is
a $224$-pixel random resized crop with a random horizontal flip; the strong
view adds \texttt{RandAugment} with three operations at magnitude nine. Images
are evaluated at a $256$-pixel resize followed by a $224$-pixel centre crop.
All three pipelines use ImageNet channel statistics.}

The benchmark of Section~\ref{sec:benchmark} and the ablations of Section~\ref{sec:ablations}
repeat each configuration over three seeds and report the mean and standard
deviation. \rev{Every standard deviation quoted in this paper is
the population value of the runs reported, with denominator $n$ rather than
$n-1$.} The pipeline is deterministic given a seed, so all
variation reported here is seed variation rather than run-to-run noise.
\rev{The last-epoch model is reported rather than a best-epoch checkpoint,
since by (P3) no target labels are available for model selection in a genuine
source-free deployment; this follows the standard protocol ~\citep{liang2020shot} and
avoids an optimistic bias that would otherwise inflate every method in the
comparison.} Reporting the last epoch is safe, because
training is stable
by then: on the VOC$\leftrightarrow$Clipart runs target mAP moves by less than
half a point over the final ten epochs, and an oracle choice of the best epoch
-- made with target labels a deployment would not have -- would add at most
$0.4$ mAP and $0.2$ on average.

\rev{Two choices in the method of Section~\ref{sec:method} are retained for
comparability and economy rather than for the gain they bring, and
Section~\ref{sec:ablations} measures both as close to neutral. The classifier
head is frozen, which keeps the parameter-matched reading of the benchmark
available, since Section~\ref{sec:trainclf} reports every baseline under the
same restriction; the comparison can then be read either with each method at
the parameters its own publication adapts, or with all of them
adapting the representation alone. Prototypes are static.}

Crucially, \emph{a single hyper-parameter configuration is used for every
adaptation task}: no per-dataset tuning is performed anywhere in this study,
for \ours{} or for any baseline. The loss weights are therefore not optimized
for any individual task, and the figures reported below are not a tuned
optimum; a per-task search could only improve on them. Method-specific hyper-parameters -- loss weights, neighbourhood sizes,
pseudo-label rounds and the like -- follow each baseline's original
publication, and the epoch budget does likewise, which for SHOT means 15
epochs and for the other three 45. Imposing a common budget would place three
of the four baselines outside their published operating point, so the
published budgets are kept. Which parameters each method adapts also follows its
publication: SHOT freezes its classifier head, while UCON, NVC-LLN and
AC-SFDA adapt theirs alongside the representation. \rev{The protocol is nonetheless only partly harmonized with those
publications, and the departures are stated rather than glossed. The largest
is that} optimization settings are held common, namely
SGD with a cosine schedule at learning rate $10^{-3}$ for UCON, NVC-LLN and
AC-SFDA, whereas their publications scale the backbone rate below the head
rate by factors of two to ten, and AC-SFDA uses Adam. \rev{That rate is the
one their own publications assign to the classifier and bottleneck, applied to
the backbone as well; \ref{app:baselines} gives the reasoning and what
it costs. \rev{Per-method departures -- UCON's uncertainty criterion and
repulsion decay, NVC-LLN's augmentation arrangement, AC-SFDA's normalization
--- are recorded in \ref{app:baselines}. The baselines should therefore be
read as multi-label extensions under a partly harmonized protocol rather than
as each method at its published operating point.}}

\subsection{\rev{Extending} single-label SFDA baselines}\label{sec:baselines}

\rev{To the best of our knowledge, no multi-label SFDA baseline exists for images. Of the two multi-label
source-free methods in Section~\ref{sec:related}, NOTELA is evaluated on
bioacoustic recordings and its image benchmarks are single-label; it is run
here at its published parameter set and epoch budget, under the same common
optimization setting as the extended baselines. CoWA addresses test-time adaptation rather than this setting, and the difference is not one an extension could absorb: its single streaming pass and its restriction to the normalization parameters are constitutive of the method, so granting it the full target set over repeated epochs would yield a different method. Four representative single-label methods are
therefore extended.}
Changes common to all four are: softmax replaced by per-class sigmoids;
categorical cross entropy replaced by binary cross entropy; and the information
maximization diversity term recomputed per class as in
Section~\ref{sec:method}. One further change proved necessary twice
independently. Both UCON and NVC-LLN promote prediction dispersion by
minimizing the dot product between the predicted distributions of different
samples in a batch, which is well defined for softmax outputs but degenerate
for independent sigmoids, since that product is minimized trivially by driving
every probability to zero. In both cases the extension $L_{2}$-normalizes the sigmoid
vectors before taking the product, penalizing the cosine similarity between
predictions rather than their magnitude. That the same repair is required in
two methods of different lineage is a concrete instance of \rev{(P1)}.

\rev{One policy governs all four extensions and is stated here because it
decides several of the choices in \ref{app:baselines}: where a mechanism has no
multi-label counterpart, the most literal per-class reading is taken rather
than a replacement designed for this study. The alternative is worse than it
looks. A designed substitute -- an adaptive confidence criterion in place of
UCON's logit ratio, say, or a multi-label pseudo-labelling rule in place of
SHOT's nearest-centroid assignment -- would make the baseline partly a new
method, and the comparison would then measure that method rather than how far
the published one carries.}

\subsection{Evaluation}\label{sec:evaluation}

\rev{Mean average precision (mAP) is reported together with label-based
(macro-averaged) and overall (micro-averaged) precision, recall and $F_{1}$ at
a threshold of $0.5$, the standard metric set for multi-label
classification~\citep{ddamlic}. mAP is threshold-free and is the primary
metric. A class with no positive annotation in an evaluation split would be
excluded from all three metrics, though this does not arise: every class is
present in every split used here. A class that is present but never predicted
contributes zero rather than being dropped.}

\rev{Each table additionally carries an unweighted mean over the five
adaptation tasks. The benchmark asks whether a method helps across domain gaps
rather than on a favourable one, and a single summary is the only form in which
that question has a direct answer; reporting a mean over a fixed set of
transfer tasks is also the convention in the single-label SFDA literature these
methods are drawn from~\citep{liang2020shot,nrc}. It is never used alone: every
claim drawn from it is accompanied by the per-task figures, and
Section~\ref{sec:benchmark} reports a case -- SHOT -- in which the mean is
uninformative precisely because gains and losses on different tasks cancel.
Means for the source-accessing block are computed here from the per-task
figures those methods publish.}

\rev{\subsection{Experiments}}

\rev{Six experiments follow, one per subsection of
Section~\ref{sec:results}.}
\begin{enumerate}
\item \rev{\emph{Benchmark} (Section~\ref{sec:benchmark}). Every method on all
      five adaptation tasks under the protocol of
      Section~\ref{sec:setup}, against the unadapted source model.}
\item \rev{\emph{Frozen-head protocol} (Section~\ref{sec:trainclf}). The same
      comparison with every method restricted to adapting the representation,
      to establish whether the ranking depends on how many parameters each
      method adapts.}
\item \rev{\emph{Component ablation} (Section~\ref{sec:ablations}). Each loss
      term, the prototype initialization and two design choices removed or
      replaced in turn, over three seeds on the two
      VOC$\leftrightarrow$Clipart directions.}
\item \rev{\emph{Alignment diagnostic} (Section~\ref{sec:contrastive}). The
      prototype contrastive term scored against the withheld ground truth and
      against the model's own pseudo-labels, to separate genuine feature-space
      alignment from self-reinforcement.}
\item \rev{\emph{Qualitative analysis} (Section~\ref{sec:qualitative}). The
      size and composition of the predicted label sets, which the per-class
      metrics of Section~\ref{sec:evaluation} cannot express.}
\item \rev{\emph{Negative result} (Section~\ref{sec:negative}).
      Confidence-based pseudo-label filtering at the level of individual
      (sample, class) entries, reported because it does not improve on
      retaining every pseudo-label.}
\end{enumerate}

\section{Results and Analysis}\label{sec:results}
\subsection{Benchmark}\label{sec:benchmark}

Table~\ref{tbl:main} reports mAP for all five adaptation tasks with a
ResNet-101 backbone, and Table~\ref{tbl:f1} the corresponding $F_{1}$ scores.

\begin{table*}[t]
\caption{Target-domain mAP (\%) on the five multi-label SFDA tasks, ResNet-101.
Source-only is the unadapted source model. Baselines marked $^{\ast}$ are
single-label SFDA methods extended to multi-label as described in
Section~\ref{sec:baselines}; \rev{NOTELA is natively multi-label and is run
without extension}. \rev{Entries are the mean and standard deviation over
three independent runs (seeds $1$, $2$ and $9$). Source-only performs no
target training and is therefore deterministic; Mean teacher + BCE is a single
run.} Among source-free methods, best in bold and second best underlined;
\rev{where a method's per-seed difference from the leader changes sign, both
are shown in bold}.}\label{tbl:main}
\small
\begin{tabular*}{\tblwidth}{@{}LLCCCCCC@{}}
\toprule
Method & Venue & UCM$\rightarrow$AID & AID$\rightarrow$UCM & VOC$\rightarrow$Clip.
       & Clip.$\rightarrow$VOC & City$\rightarrow$Foggy & Avg. \\
\midrule
Source-only & -- & 52.32 & \underline{54.22} & 39.77 & 55.52 & \underline{55.88} & 51.54 \\
\rev{Mean teacher + BCE$^{\ddagger}$} & \rev{--} & \underline{56.05} & 52.02 & 38.25 & 60.85 & 55.44 & 52.52 \\
SHOT$^{\ast}$ & ICML 2020 & 53.59\sd{0.08} & 44.65\sd{0.28} & \underline{45.11\sd{0.30}} & 61.64\sd{0.08} & 52.71\sd{0.14} & 51.54\sd{0.07} \\
\rev{NOTELA} & \rev{ICML 2023} & 51.73\sd{0.15} & 51.60\sd{0.20} & 39.70\sd{0.14} & 60.73\sd{0.04} & \textbf{56.84\sd{0.05}} & 52.12\sd{0.05} \\
UCON$^{\ast}$ & ICLR 2025 & 36.84\sd{4.97} & 21.45\sd{0.94} & 14.86\sd{2.64} & 18.38\sd{3.99} & 49.31\sd{0.99} & 28.17\sd{1.22} \\
NVC-LLN$^{\ast}$ & TPAMI 2025 & \textbf{56.52\sd{0.43}} & 53.60\sd{0.67} & 40.92\sd{0.40} & \textbf{68.52\sd{0.19}} & 51.52\sd{0.32} & \underline{54.22\sd{0.19}} \\
AC-SFDA$^{\ast}$ & ESWA 2026 & 54.06\sd{0.33} & 53.15\sd{1.18} & 40.25\sd{1.91} & 60.51\sd{0.05} & 55.50\sd{0.44} & 52.69\sd{0.29} \\
\ours{} & ours & \textbf{56.21\sd{0.82}} & \textbf{57.20\sd{1.63}} & \textbf{45.28\sd{0.40}} & \underline{63.15\sd{0.59}} & \textbf{56.71\sd{0.30}} & \textbf{55.71\sd{0.57}} \\
\midrule
DA-MAIC$^{\dagger}$ & TGRS 2022 & 50.50 & 54.40 & 62.30 & 74.30 & 61.90 & 60.68 \\
DALN$^{\dagger}$ & CVPR 2022 & 53.20 & 53.10 & 45.00 & 66.70 & 54.80 & 54.56 \\
DDA-MLIC$^{\dagger}$ & ESWA 2026 & 54.90 & 63.20 & 61.40 & 77.00 & 62.30 & 63.76 \\
DDA-MLIC + DeepEM$^{\dagger}$ & ESWA 2026 & 56.20 & 60.50 & 62.00 & 82.80 & 63.20 & 64.94 \\
\bottomrule
\end{tabular*}

\vspace{2pt}
\footnotesize $^{\ast}$Single-label SFDA method extended to multi-label for
this study. \rev{$^{\ddagger}$Plain multi-label mean teacher trained with
binary cross entropy: the framework of Section~\ref{sec:method} with all three
proposed terms removed.} $^{\dagger}$Multi-label UDA methods with full source access; \rev{figures as
published by their authors, not rerun here}. Not source-free, and not directly
comparable.
\end{table*}

\begin{table*}[t]
\caption{Label-based ($F_{1}^{L}$) and overall ($F_{1}^{O}$) $F_{1}$ (\%) at
threshold $0.5$ (ResNet-101) on all five adaptation tasks. \rev{Mean and
standard deviation over seeds $1$, $2$ and $9$, with the same single-run
exceptions as Table~\ref{tbl:main}.} Among source-free
methods, best in bold and second best underlined\rev{, under the tie rule of
Table~\ref{tbl:main}}; the source-accessing block is
excluded from both rankings. Markers as in Table~\ref{tbl:main}.}\label{tbl:f1}
\small
\begin{tabular*}{\tblwidth}{@{}LCCCCCC@{}}
\toprule
Method & UCM$\rightarrow$AID & AID$\rightarrow$UCM & VOC$\rightarrow$Clip.
       & Clip.$\rightarrow$VOC & City$\rightarrow$Foggy & Avg. \\
\midrule
\multicolumn{7}{@{}l}{\emph{Label-based $F_{1}^{L}$}}\\
Source-only & 38.27 & \underline{50.78} & 25.91 & 41.02 & 45.60 & 40.32 \\
\rev{Mean teacher + BCE$^{\ddagger}$} & 37.69 & 47.84 & 27.13 & 49.46 & 45.09 & 41.44 \\
SHOT$^{\ast}$ & \underline{42.89\sd{0.21}} & 46.98\sd{0.32} & \underline{36.79\sd{0.40}} & 52.34\sd{0.08} & 48.36\sd{0.12} & \underline{45.47\sd{0.16}} \\
\rev{NOTELA} & 35.14\sd{0.25} & 46.88\sd{0.37} & 25.20\sd{0.23} & 49.64\sd{0.27} & \underline{51.13\sd{0.17}} & 41.60\sd{0.17} \\
UCON$^{\ast}$ & 23.96\sd{4.31} & 19.44\sd{1.00} & 3.04\sd{0.60} & 16.00\sd{4.43} & 37.49\sd{6.00} & 19.99\sd{0.76} \\
NVC-LLN$^{\ast}$ & 27.00\sd{0.17} & 33.16\sd{1.08} & 31.51\sd{0.26} & \textbf{60.28\sd{0.06}} & 11.74\sd{0.27} & 32.74\sd{0.20} \\
AC-SFDA$^{\ast}$ & 36.10\sd{0.58} & 45.27\sd{0.89} & 19.55\sd{1.10} & 39.19\sd{0.46} & 39.93\sd{0.66} & 36.01\sd{0.17} \\
\ours{} & \textbf{44.93\sd{0.65}} & \textbf{52.28\sd{0.48}} & \textbf{37.72\sd{0.87}} & \underline{53.59\sd{0.66}} & \textbf{54.50\sd{0.40}} & \textbf{48.60\sd{0.35}} \\
\cmidrule(l){1-7}
DA-MAIC$^{\dagger}$ & 29.00 & 38.60 & 51.60 & 63.00 & 42.70 & 44.98 \\
DALN$^{\dagger}$ & 32.70 & 36.70 & 32.60 & 31.70 & 25.40 & 31.82 \\
DDA-MLIC$^{\dagger}$ & 35.50 & 55.10 & 39.40 & 38.20 & 48.90 & 43.42 \\
DDA-MLIC + DeepEM$^{\dagger}$ & 26.50 & 48.80 & 34.60 & 65.80 & 45.40 & 44.22 \\
\midrule
\multicolumn{7}{@{}l}{\emph{Overall $F_{1}^{O}$}}\\
Source-only & \underline{67.86} & \textbf{69.51} & 37.19 & 58.53 & 71.09 & 60.84 \\
\rev{Mean teacher + BCE$^{\ddagger}$} & 59.83 & 67.99 & 44.84 & \underline{63.50} & \textbf{76.76} & 62.58 \\
SHOT$^{\ast}$ & 61.67\sd{0.47} & 62.78\sd{0.23} & \underline{51.96\sd{0.42}} & 58.86\sd{0.18} & 61.98\sd{0.27} & 59.45\sd{0.18} \\
\rev{NOTELA} & 62.13\sd{0.38} & 67.32\sd{0.11} & 45.92\sd{0.25} & \textbf{63.67\sd{0.12}} & 74.75\sd{0.28} & \underline{62.76\sd{0.09}} \\
UCON$^{\ast}$ & 42.14\sd{6.50} & 37.23\sd{2.98} & 20.45\sd{4.25} & 24.27\sd{2.19} & 65.64\sd{5.68} & 37.95\sd{1.22} \\
NVC-LLN$^{\ast}$ & 21.04\sd{0.13} & 25.58\sd{0.72} & 29.39\sd{0.20} & 55.68\sd{0.22} & 12.59\sd{0.57} & 28.86\sd{0.28} \\
AC-SFDA$^{\ast}$ & \textbf{68.67\sd{0.28}} & \underline{68.79\sd{0.93}} & 41.79\sd{0.92} & 57.32\sd{0.57} & \underline{76.26\sd{0.72}} & 62.57\sd{0.26} \\
\ours{} & 65.02\sd{0.25} & 68.10\sd{0.42} & \textbf{55.06\sd{0.60}} & \textbf{63.01\sd{0.49}} & 75.11\sd{0.05} & \textbf{65.26\sd{0.31}} \\
\cmidrule(l){1-7}
DA-MAIC$^{\dagger}$ & 50.80 & 67.90 & 63.20 & 68.30 & 71.70 & 64.38 \\
DALN$^{\dagger}$ & 54.80 & 60.60 & 36.40 & 30.00 & 49.20 & 46.20 \\
DDA-MLIC$^{\dagger}$ & 55.30 & 69.20 & 48.80 & 50.40 & 76.00 & 59.94 \\
DDA-MLIC + DeepEM$^{\dagger}$ & 47.60 & 68.20 & 40.00 & 70.00 & 75.50 & 60.26 \\
\bottomrule
\end{tabular*}
\end{table*}

Three observations stand out. First, among source-free methods \ours{} obtains
the best average on every metric and the best or tied-best mAP on \rev{four} of the five tasks,
improving on the strongest \rev{extended} baseline \rev{by mAP} (NVC-LLN) by $1.49$
mAP and on the strongest by label-based $F_{1}$ (SHOT) by
$3.13$ points. \rev{It is passed on three tasks, one of which is a genuine
loss and two of which are not. The loss is} Clipart$\rightarrow$VOC, where
NVC-LLN \rev{leads by $5.37 \pm 0.56$ mAP, the same sign in all three seeds}.
That reversal is informative, and Section~\ref{sec:trainclf} shows it depends
on NVC-LLN adapting its classifier head: with every method restricted to a
frozen head \rev{the ordering is restored}. \rev{The other two are statistical
ties. NVC-LLN leads by $0.32 \pm 0.68$ on UCM$\rightarrow$AID and NOTELA by
$0.13 \pm 0.35$ on Cityscapes$\rightarrow$Foggy, and in both the sign of the
per-seed difference changes across the three seeds, so neither ordering is
established by these runs -- which is why Table~\ref{tbl:main} marks both
entries in bold. Cityscapes$\rightarrow$Foggy is also the task with the
smallest appearance shift in the benchmark.}

Second, and more consequential for how this problem should be framed, \rev{extending}
a single-label SFDA method to multi-label data does not reliably beat doing
nothing at all. Two of the four clear the unadapted source model on average,
NVC-LLN by $2.67$ average mAP and AC-SFDA by $1.15$; UCON collapses ($-23.37$), and
SHOT is the clearest case of the pattern: its gains on three tasks and losses
on the other two cancel almost exactly, leaving it level with source-only
($51.54$ versus $51.54$). On two of the five tasks, AID$\rightarrow$UCM and
Cityscapes$\rightarrow$Foggy, no \rev{extended} baseline beats source-only at all.
\ours{} is the only method that improves on source-only on every task.
\rev{The narrowest of those margins is Cityscapes$\rightarrow$Foggy, at $0.83$
mAP on average and $0.54$ in the weakest of the three seeds: of the same order
as the seed-to-seed spread, but positive in every seed.} Part of
SHOT's behaviour is structural: its centroid assignment emits a single positive
label per image (Section~\ref{sec:baselines}), yet because evaluation reads the
model's per-class scores rather than its pseudo-labels, a single-hot training
signal still leaves the ranking usable. The damage surfaces not as uniform
collapse but as unreliability, which is harder to detect and more dangerous in
deployment, since a practitioner cannot tell in advance which side of the ledger
a given domain pair will land on. UCON degrades furthest, reaching $3.04$
$F_{1}^{L}$ on VOC$\rightarrow$Clipart. \rev{Its uncertainty control has no
per-class counterpart and had to be replaced for this study;
\ref{app:baselines} shows that the replacement is not what the collapse rests
on.}

\rev{The second row of Table~\ref{tbl:main} bears on how that should be read.
A plain multi-label mean teacher trained with binary cross entropy -- the
framework of Section~\ref{sec:method} with all three proposed terms removed --
averages $52.52$ mAP. That is $0.98$ above source-only and among the baselines, 
between NOTELA at $52.12$ and AC-SFDA at $52.69$, while
\ours{} leads it by $3.19$. A framework carrying no single-label mechanism at
all therefore performs about as well as the methods that carry one, which is
evidence that the extensions are not what holds those methods back. It remains
below \ours{} on all of the five tasks and below source-only on three, so it
is a reasonable baseline rather than a competitive method.}

\rev{Third, two further gaps separate the methods, and neither is visible in
mAP. The first is between how well a method \emph{orders} its scores and how
well those scores sit against the decision threshold. NVC-LLN is the extreme
case: the strongest baseline by mAP ($54.22$) and the weakest of all by overall
$F_{1}$ ($28.86$). All $F_{1}$ figures are taken at $0.5$, the
operating point a deployed model has: mAP summarizes the ranking across
thresholds but does not select one, and the threshold would normally be fixed
on a labelled validation split, which (P3) does not provide. The two metrics
together therefore measure ranking quality and performance at one fixed
operating point. NVC-LLN ranks well and scores poorly there; which property of
its score distribution produces that gap is not something these two metrics
settle.}

\rev{The second gap is between frequent and rare classes, which the two
$F_{1}$ variants separate by construction. AC-SFDA shows a $26.56$-point spread
between its overall and label-based averages,
consistent with strong performance on frequent classes and weak performance on
rare ones -- though at a single threshold that reading cannot be fully
separated from the fixed-threshold effect above.} \ours{} is the only method to
lead on both halves at once, so its advantage does not
come from trading \rev{rare-class performance for frequent-class
performance}. This is
exactly the failure mode the label-based average is designed to expose, and it
is invisible in the single-label setting.

Finally, the source-accessing multi-label UDA methods in the lower block
indicate what is achievable once the constraint is lifted. \rev{Their figures
are reproduced as published rather than rerun under the protocol of
Section~\ref{sec:setup}, so source training, splits, preprocessing and
optimization all differ: the block is an external reference, not a matched
comparison.} The block spans both families distinguished by~\citet{ddamlic}: DA-MAIC uses a
domain discriminator, while DALN and DDA-MLIC dispense with one. DDA-MLIC and
its DeepEM variant lead \ours{} by $8.05$ and $9.23$ mAP on average, as expected
given full source supervision, and their advantage is widest on the two
VOC/Clipart directions ($14$--$20$ points), where the appearance shift is
largest. The
picture inverts at the operating threshold: \ours{} attains the highest
label-based and overall $F_{1}$ of any method in the table, source-free or not,
exceeding the best source-accessing method on either measure (DA-MAIC) by
$3.62$ $F_{1}^{L}$ and $0.88$ $F_{1}^{O}$ on average. \rev{The two orderings
therefore disagree between the metrics; what produces that is not something an
uncontrolled comparison can settle.} This is an average effect rather than a uniform one:
some UDA method surpasses \ours{} on $F_{1}^{L}$ on three of the five tasks,
and \ours{} leads on average through consistency rather than dominance.
\rev{DALN~\citep{daln} averages $1.15$ mAP below \ours{} despite training with
the source data throughout, so the block is not uniformly ahead either.}

\subsection{Restricting every method to a frozen classifier head}\label{sec:trainclf}

Table~\ref{tbl:main} follows each baseline's publication on which parameters
it adapts, so SHOT adapts its representation only while UCON, NVC-LLN and
AC-SFDA also adapt their classifier heads. That is the fairest reading of each method, but it
means the comparison is not parameter-matched. This section removes that
difference by freezing the head for all three, changing nothing else, and
re-running every task.

\begin{table*}[t]
\caption{Target-domain mAP (\%) for the three baselines whose published
configurations adapt the classifier head, as reported in
Table~\ref{tbl:main} and with the head instead frozen. \rev{The left block
averages over the five adaptation tasks, the right block
gives $\Delta$ task by task.} Single run per \rev{task} at
the fixed seed of Section~\ref{sec:setup}.}\label{tbl:trainclf}
\small
\begin{tabular*}{\tblwidth}{@{}LCCCCCCCC@{}}
\toprule
& \multicolumn{3}{c}{Average over five tasks} & \multicolumn{5}{c}{\rev{Per-task $\Delta$}} \\
\cmidrule(lr){2-4}\cmidrule(lr){5-9}
Method & Published & Head frozen & $\Delta$
 & UCM$\rightarrow$AID & AID$\rightarrow$UCM & VOC$\rightarrow$Clip.
 & Clip.$\rightarrow$VOC & City$\rightarrow$Foggy \\
\midrule
UCON    & 29.51 & 29.51 & $-0.00$ & \rev{$-3.23$} & \rev{$+7.54$} & \rev{$-3.95$} & \rev{$+0.38$} & \rev{$-0.72$} \\
NVC-LLN & 54.13 & 50.48 & $-3.65$ & \rev{$-4.60$} & \rev{$-8.33$} & \rev{$+1.89$} & \rev{$-5.88$} & \rev{$-1.34$} \\
AC-SFDA & 52.76 & 53.12 & $+0.36$ & \rev{$+1.76$} & \rev{$-0.22$} & \rev{$-0.70$} & \rev{$+0.47$} & \rev{$+0.56$} \\
\bottomrule
\end{tabular*}
\end{table*}

Table~\ref{tbl:trainclf} shows that the ranking is unchanged. \ours{} leads
the strongest baseline by $1.49$ mAP \rev{on average across the five tasks}
when each method runs as published. \rev{At the fixed seed of this experiment
that lead is $2.05$, and it widens to $3.06$} when every method is
restricted to adapting its representation only. The conclusion therefore does
not rest on which convention is adopted, which is the property that matters: a
comparison configured to favour the proposed method would show the ordering
changing once the convention was altered. Under the frozen-head protocol
\ours{} also regains Clipart$\rightarrow$VOC, since NVC-LLN falls from $68.42$
to $62.54$ \rev{at that seed}.

The three baselines respond very differently, and the pattern is the same one
that organizes Section~\ref{sec:benchmark}. UCON is unmoved to two decimal
places on average, \rev{but Table~\ref{tbl:trainclf} shows that this
conceals large changes of opposite sign, from $-3.95$ to $+7.54$ across the
five tasks, which happen to cancel to within a hundredth of a point. That the
head matters task by task and yet leaves the average where it was is} strong
evidence that its collapse is
intrinsic to its uncertainty mechanism rather than a
consequence of how many parameters it adapts. AC-SFDA moves by $0.36$, within
the range of seed variation reported in Section~\ref{sec:ablations}; it
already maintains an offline teacher as its pseudo-label source, so an
adaptable head adds little. Only NVC-LLN depends materially on the choice, and
it is the baseline whose objective is furthest from any single-label
assumption: its losses are per-class binary cross entropies between a sample's
predictions and those of its feature-space neighbours, so nothing in it
\rev{depends on (P1)}. It is the one baseline for which the
parameter-matched protocol is genuinely restrictive, and the only one that
changes rank between the two conventions.

\subsection{Ablation of design choices}\label{sec:ablations}

\begin{figure*}[t]
  \centering
  \includegraphics[width=0.9\textwidth]{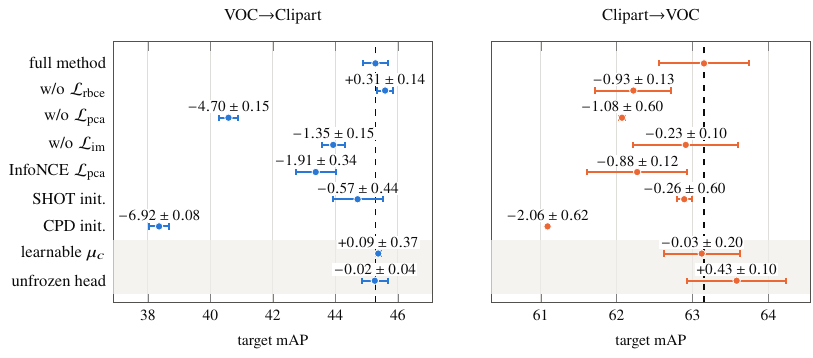}
  \caption{Component ablation on the two VOC$\leftrightarrow$Clipart
  directions. Each point is the mean mAP (\%) over
  three seeds and the bar spans $\pm 1$ standard deviation.
  The dashed line marks the full method. \rev{Shaded rows are the design choices not corresponding to the effect of a contribution.
  \emph{InfoNCE $\mathcal{L}_{\mathrm{pca}}$} replaces the prototype
  contrastive term with the conventional multi-positive form, normalized over
  the prototypes for each sample rather than over the samples for each
  prototype.} The
  value above each point is the mean
  \rev{\emph{per-seed difference}}: the listed configuration minus the full
  method \rev{on the same seed, averaged over the three seeds}, $\pm$ its
  standard deviation. A negative value
  indicates that the listed change degrades performance. The full method itself reaches
  $45.28 \pm 0.40$ and $63.15 \pm 0.59$ mAP on the two directions.}\label{fig:ablation}
\end{figure*}

Figure~\ref{fig:ablation} isolates the contribution of each component. The
ablation is run on the two VOC$\leftrightarrow$Clipart directions.
\rev{Although the two directions use the same pair of datasets, the source
model performs very differently on them}: it reaches only
$39.77$ mAP on Clipart and $55.52$ on VOC, the worst and among the best
source-to-target transfers in the benchmark. Each configuration is repeated
over three random seeds.

Prototype contrastive alignment \rev{($\mathcal{L}_{\text{pca}}$) is the
largest contributor among the three loss terms, by a wide margin}.
Removing it costs $4.70$ mAP on VOC$\rightarrow$Clipart and $1.08$ on
Clipart$\rightarrow$VOC. \rev{On VOC$\rightarrow$Clipart the per-seed
difference is $4.70 \pm 0.15$, against standard deviations of $0.40$ and $0.31$
for the two configurations measured separately. The per-seed form is used
because it carries a spread for the difference itself, which the two individual
standard deviations do not give: a seed displaces both arms in the same
direction, so that displacement cancels in the difference and the gap is known
nearly three times more precisely than either level is. The same holds for
every entry of Figure~\ref{fig:ablation}, which is why the per-seed differences
rather than the individual standard deviations are quoted throughout.}

\rev{Replacing that term with the conventional multi-positive prototype
InfoNCE, normalized over the $C$ prototypes for each sample rather than over
the $B$ samples for each prototype, costs $1.91 \pm 0.34$ mAP on
VOC$\rightarrow$Clipart and $0.88 \pm 0.12$ on Clipart$\rightarrow$VOC, and is
negative on all six per-seed differences. The normalization axis of
Section~\ref{sec:contrastive-loss} is therefore a measured choice rather than
only a principled one. The two forms differ in a second respect, which the
comparison does not separate: normalizing per sample cancels the
$\lVert\tilde{\bvec{y}}_{j}\rVert_{2}$ in $w_{cj}$, so the conventional form
also loses the label-density weighting.}

Information maximization \rev{($\mathcal{L}_{\text{im}}$)} contributes
consistently but far less: $1.35 \pm 0.15$
on VOC$\rightarrow$Clipart and $0.23 \pm 0.10$ on Clipart$\rightarrow$VOC,
positive in all six \rev{per-seed differences}. \rev{It therefore helps in
spite of the $\bar{q}_{c}=0.5$ target that Section~\ref{sec:method} identifies
as wrong for rare classes, rather than because that mismatch is harmless.}

The effect of removing the reverse term of the symmetric objective \rev{($\mathcal{L}_{\text{rbce}}$, leaving $\mathcal{L}_{\text{bce}}$ alone)} is small and reverses sign
between the two directions: \rev{the full method} is ahead by $0.93$ mAP on
Clipart$\rightarrow$VOC and behind by $0.31$ on VOC$\rightarrow$Clipart,
\rev{each ordering holding on all three seeds. Averaging the two directions
within a seed leaves the full method ahead on all three, by $0.31$.} The term
is therefore retained, but as a modest and
direction-dependent contribution.

\rev{The prototype initialization of Section~\ref{sec:protoinit} is the
contribution the evidence supports most strongly.} Applying the
weighting of~\citet{cpd} verbatim, with exponential weights normalized over the
whole batch and no threshold \rev{(\emph{CPD init.} in
Figure~\ref{fig:ablation})}, costs $4.50$ mAP \rev{on average across the two
directions. This is the largest effect measured anywhere in this study, and on
VOC$\rightarrow$Clipart it is also the most stable of the large effects.} It
confirms the analysis of
Section~\ref{sec:protoinit}: a prototype drawing
most of its mass from images that do not contain its class is a misleading anchor, and \rev{$\mathcal{L}_{\text{pca}}$} then pulls features towards
it. That the damage exceeds \rev{removing $\mathcal{L}_{\text{pca}}$}
altogether \rev{-- $4.50$ against $2.89$ averaged over the two directions --}
is the
clearest evidence in the paper that alignment to a corrupted anchor is worse
than no alignment, which is the same conclusion reached from a different
direction by the diagnostic of Section~\ref{sec:contrastive}.

A milder variant of the same choice moves results far less. Replacing the confidence-weighted scheme of
Section~\ref{sec:protoinit} with the SHOT-style global soft-weighted centroid
of~\citet{liang2020shot}, refined over three rounds \rev{(\emph{SHOT init.})},
costs $0.42$ mAP \rev{on average}. The evidence is stronger on the first direction, where the full
method is ahead in all three seeds, than on the second, where the sign of the
per-seed difference varies by seed and only the mean favours the chosen scheme.
The asymmetry is the same one seen for the two feature-space terms: the
initialization matters more where the source model transfers worst, which is
consistent with prototypes computed on badly shifted features being harder to
place well, and with the initialization mattering less once the target
features already resemble the source.

\rev{The remaining entries are not among the stated contributions}, and the
role of the figure there is to record what the evidence does and does not
support.

Making the prototypes learnable changes nothing measurable. The per-seed
difference is $+0.09 \pm 0.37$ on VOC$\rightarrow$Clipart and
$-0.03 \pm 0.20$ on Clipart$\rightarrow$VOC, and the sign disagrees across
seeds on both directions. \rev{The absence of an effect} is not an artifact of
the prototypes failing to move: they are optimized under the same schedule and learning rate
as the backbone, and their accumulated gradient magnitude at the end of
adaptation is roughly an order of magnitude larger than that of the early
convolutional layers. Nor is it an artifact of how $\mathcal{L}_{\text{pce}}$
is balanced against the rest of the objective; reducing its weight tenfold,
from $0.1$ to $0.01$, moves the result by $+0.01 \pm 0.06$ and
$-0.03 \pm 0.02$ on the two directions. Static prototypes are therefore
retained on grounds of parsimony\rev{: they remove a parameter group, a loss
term and a weight from the method at no measurable cost}.

Unfreezing the classifier head gives the one result in the figure that runs
against the design. It is neutral on VOC$\rightarrow$Clipart
($-0.02 \pm 0.04$) but improves Clipart$\rightarrow$VOC by $0.43 \pm 0.10$,
positive in all three seeds, for a two-task average gain of $0.21$. Freezing
the head is therefore not necessary for the method to work, and on these two
tasks it costs \rev{a small amount of target mAP}, an effect about two-thirds
the size of the reverse term's contribution. The head is nevertheless kept
frozen in every reported result, \rev{for the two reasons given in
Section~\ref{sec:setup}; the trade-off is recorded here rather than resolved,
and whether the gain persists on the remaining three tasks is not established
by these experiments}.

\subsection{When does prototype alignment actually help?}\label{sec:contrastive}

The prototype contrastive term behaves differently across tasks, and
understanding why is informative about prototype-based adaptation
in general. Table~\ref{tbl:contrastive} shows its effect with all other
components fixed, \rev{on all five adaptation tasks}.

\begin{table*}[t]
\caption{Effect of the prototype contrastive alignment term
($\lambda_{\text{pca}}\!=\!0$ versus $0.5$) \rev{on all five tasks},
ResNet-101. \rev{Mean and standard deviation over three independent runs
(seeds $1$, $2$ and $9$), with every other setting of
Section~\ref{sec:setup} held fixed, so the $\lambda_{\text{pca}}=0.5$ columns
are the \ours{} entries of Tables~\ref{tbl:main} and~\ref{tbl:f1}. Each
$\Delta$ is the mean of the three per-seed differences together with the
spread of those differences. $^{\S}$ marks a $\Delta$ whose three per-seed
differences do not all share a sign, so these runs do not establish the
direction of the effect. The diagnostic of
Figure~\ref{fig:auc} is computed from a separate single run per task that
refreshes pseudo-labels every epoch rather than every
third.}}\label{tbl:contrastive}
\small
\begin{tabular*}{\tblwidth}{@{}LCCCCCCCCC@{}}
\toprule
 & \multicolumn{3}{c}{mAP} & \multicolumn{3}{c}{$F_1^L$} & \multicolumn{3}{c}{$F_1^O$} \\
\cmidrule(lr){2-4}\cmidrule(lr){5-7}\cmidrule(lr){8-10}
Task & w/o PCA & with PCA & $\Delta$ & w/o PCA & with PCA & $\Delta$
     & w/o PCA & with PCA & $\Delta$ \\
\midrule
VOC$\rightarrow$Clipart & 40.58\sd{0.31} & 45.28\sd{0.40} & $+4.70$\sd{0.15} & 30.45\sd{0.32} & 37.72\sd{0.87} & $+7.26$\sd{0.67} & 49.38\sd{0.26} & 55.06\sd{0.60} & $+5.68$\sd{0.39} \\
Clipart$\rightarrow$VOC & 62.07\sd{0.03} & 63.15\sd{0.59} & $+1.08$\sd{0.60} & 52.84\sd{0.09} & 53.59\sd{0.66} & $+0.75^{\S}$\sd{0.61} & 64.61\sd{0.13} & 63.01\sd{0.49} & $-1.60$\sd{0.57} \\
AID$\rightarrow$UCM & 56.69\sd{0.29} & 57.20\sd{1.63} & $+0.51^{\S}$\sd{1.54} & 52.20\sd{0.19} & 52.28\sd{0.48} & $+0.08^{\S}$\sd{0.30} & 68.63\sd{0.26} & 68.10\sd{0.42} & $-0.53$\sd{0.17} \\
Cityscapes$\rightarrow$Foggy & 57.25\sd{0.31} & 56.71\sd{0.30} & $-0.54^{\S}$\sd{0.57} & 52.50\sd{0.42} & 54.50\sd{0.40} & $+2.00$\sd{0.33} & 75.24\sd{0.05} & 75.11\sd{0.05} & $-0.13$\sd{0.03} \\
UCM$\rightarrow$AID & 57.65\sd{0.28} & 56.21\sd{0.82} & $-1.44$\sd{0.55} & 46.39\sd{0.05} & 44.93\sd{0.65} & $-1.47$\sd{0.69} & 65.43\sd{0.43} & 65.02\sd{0.25} & $-0.41^{\S}$\sd{0.62} \\
\midrule
\rev{Average} & 54.85\sd{0.14} & 55.71\sd{0.57} & $+0.86$\sd{0.45} & 46.88\sd{0.14} & 48.60\sd{0.35} & $+1.72$\sd{0.23} & 64.66\sd{0.00} & 65.26\sd{0.31} & $+0.60$\sd{0.31} \\
\bottomrule
\end{tabular*}
\end{table*}

\rev{The term is worth $4.70$ mAP on VOC$\rightarrow$Clipart and costs $1.44$
on UCM$\rightarrow$AID, and across the five tasks it averages $+0.86 \pm 0.45$: it pays
on two, is neutral on two and costs on one. Its benchmark-wide contribution is
therefore far smaller than the ablation of Section~\ref{sec:ablations}
suggests, because that ablation runs on the two directions where it pays.}
A natural hypothesis is that the
alignment loss is simply redundant when the domain gap is small, but that
would predict indifference rather than a loss. An alternative explains the
sign: prototypes are derived from the model's own predictions, so pulling
features towards them may manufacture agreement that is self-generated rather
than informative, \rev{in which case the term is not only redundant but
harmful, spending its capacity on reinforcing the model's existing errors}.

To separate these explanations, a \emph{prototype-agreement diagnostic} is
introduced. At each refresh it computes, for every class, the area under the
ROC curve obtained by ranking target samples according to their cosine
similarity to that class prototype, scored twice: once against the model's own
pseudo-labels ($\mathrm{AUC}_{\text{cls}}$, measuring internal consistency) and
once against the withheld ground-truth labels ($\mathrm{AUC}_{\text{gt}}$,
measuring genuine information; used for analysis only and never for training).
A term that merely manufactures agreement raises the former without the latter.

\begin{figure}
  \centering
  \includegraphics[width=0.9\columnwidth]{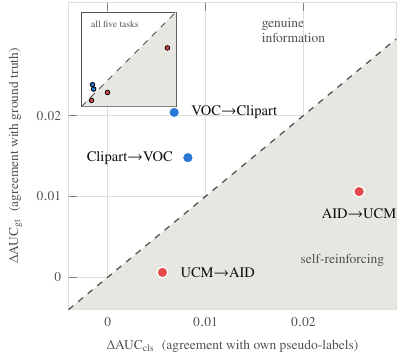}
  \caption{\rev{Prototype-agreement diagnostic. Each axis is the change, with
  minus without the contrastive term, in the AUC of ranking target samples by
  prototype similarity: against the withheld ground truth (vertical) and
  against the model's own pseudo-labels (horizontal). Points are final-epoch
  values from a single run per task. Above the diagonal the term buys more agreement with the
  labels than with the model's own beliefs. The inset places all five tasks on a single
  scale: the outlying point is Cityscapes$\rightarrow$Foggy at
  $(0.101,\,0.067)$, some $4.4$ times further from the origin than any other
  and likewise below the diagonal. The ground-truth axis is computed after
  adaptation for this analysis; it is not available during it, and is not a
  model-selection signal.}}\label{fig:auc}
\end{figure}

\rev{Figure~\ref{fig:auc} separates the two explanations by setting the two
agreements against each other: what distinguishes the tasks is how
$\mathrm{AUC}_{\text{gt}}$ moves relative to
$\mathrm{AUC}_{\text{cls}}$.}
\rev{At the final epoch, ground-truth agreement outgains pseudo-label
agreement on both VOC$\leftrightarrow$Clipart directions and on none of
AID$\rightarrow$UCM, UCM$\rightarrow$AID or Cityscapes$\rightarrow$Foggy. The
two groups do not overlap, and they are exactly the two tasks on which the
term pays and the three on which it does not. Each task is a single run, so
this is an exploratory reading rather than a test: it shows the diagnostic
tracking the sign of the term's effect, not that one causes the other.}

Two readings follow. First, on the VOC$\leftrightarrow$Clipart
directions the gains are accompanied by \rev{a larger rise in agreement with
the withheld labels than with the model's own predictions, which is consistent
with genuine feature-space reorganization rather than self-confirmation. With a
single run per task the diagnostic supports that reading rather than
establishing it}. Second, prototype
alignment is not a free addition. Where the source representation already
clusters the target data adequately -- UCM and AID share well-separated scene categories, whereas VOC and Clipart do not
-- there is little genuine structure left for the prototypes to contribute,
and the term spends its influence amplifying the model's own predictions
instead, at a measurable cost. \rev{The harm survives a change of refresh
schedule: $-2.33$ mAP under the every-epoch schedule these diagnostic runs use
and $-1.44 \pm 0.55$ under the three-epoch schedule of
Table~\ref{tbl:contrastive}.}


\subsection{Qualitative analysis}\label{sec:qualitative}

The metrics of Section~\ref{sec:benchmark} score each class independently and
therefore say nothing about the \emph{shape} of the predicted label sets. Two
structural properties are examined here on VOC$\rightarrow$Clipart.

\begin{figure}
  \centering
  \includegraphics[width=\columnwidth]{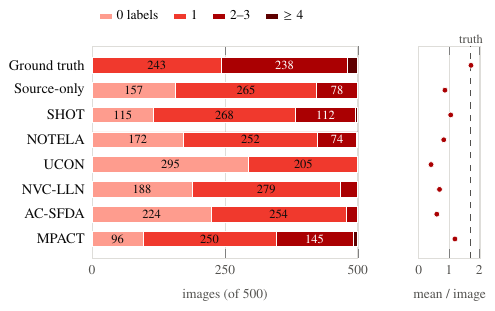}
  \caption{Number of positive labels predicted per image at threshold $0.5$ on
  the $500$ Clipart validation images, VOC$\rightarrow$Clipart. Left: how the
  $500$ images divide by predicted label-set size, with counts printed in every
  segment wide enough to carry one. Right: the mean number of labels predicted
  per image, against the ground-truth mean
  (dashed).}\label{fig:labelcount}
\end{figure}

Figure~\ref{fig:labelcount} reports how many labels each method actually emits.
The most striking entry is UCON, which predicts two or more labels on
\emph{none} of the $500$ images and nothing at all on $295$ of them. It effectively reverted to single-label prediction, which is the behaviour its
uncertainty mechanism implicitly assumes and the clearest available
confirmation of the account given in Section~\ref{sec:benchmark}. The pattern
is not confined to UCON. Every method under-predicts relative to the ground
truth mean of $1.73$ labels per image, and every method leaves a substantial
number of images with no prediction at all. \rev{NOTELA, the one method in the
comparison built for multi-label data rather than extended to it,
under-predicts more than the unadapted source model ($0.83$ against $0.87$
labels per image), so the shortfall is not an artefact of the extensions.}
\ours{} under-predicts least, at
$1.20$ labels per image with $96$ empty predictions. \rev{Three methods reach
four labels on any image -- \ours{} and SHOT on five images each, NOTELA on
two -- and \ours{} alone reaches five, on four; none of the other four exceeds
three labels.} This is
\rev{the $\bar{q}_{c}=0.5$ diversity target seen directly: an objective that
drives every class's mean predicted probability to $0.5$ still leaves every
model systematically reluctant to assert the rare ones}.

\begin{table}[t]
\caption{Label-set recovery on VOC$\rightarrow$Clipart. \emph{Exact} counts
images whose predicted label set matches the ground truth set exactly;
\emph{partial} counts those recovering at least one true label; \emph{empty}
counts those predicting nothing. \rev{The columns are not disjoint: every
exact recovery is also a partial one, so \emph{partial} includes
\emph{exact}. An image that is neither partial nor empty predicted labels, but
none of the correct ones.}}\label{tbl:setrecovery}
\small
\begin{tabular*}{\tblwidth}{@{}LCCC@{}}
\toprule
Method & Exact & Partial & Empty \\
\midrule
\multicolumn{4}{@{}l}{\emph{Images carrying two labels} ($n=171$)}\\
Source-only & 15 & 74  & 50 \\
SHOT        & 23 & 117 & 36 \\
\rev{NOTELA} & \rev{16} & \rev{103} & \rev{50} \\
UCON        & 0  & 55  & 99 \\
NVC-LLN     & 3  & 61  & 60 \\
AC-SFDA     & 8  & 98  & 68 \\
\ours{} (ours) & \textbf{35} & \textbf{127} & \textbf{24} \\
\midrule
\multicolumn{4}{@{}l}{\emph{Images carrying three or more labels} ($n=86$)}\\
Source-only & 0 & 59 & 15 \\
SHOT        & 0 & 64 & \textbf{10} \\
\rev{NOTELA} & \rev{0} & \rev{50} & \rev{23} \\
UCON        & 0 & 31 & 34 \\
NVC-LLN     & 0 & \rev{25} & \rev{40} \\
AC-SFDA     & 0 & 43 & 33 \\
\ours{} (ours) & \textbf{3} & \textbf{65} & \textbf{10} \\
\bottomrule
\end{tabular*}
\end{table}

Table~\ref{tbl:setrecovery} scores the predicted set as a whole, a criterion
the per-class metrics cannot express. On the modal case of exactly two labels
\ours{} recovers the complete set on $35$ images against $23$ for the next best
method, and leaves fewest images empty. On the label-dense images it is the
only method to recover a complete set at all, and does so three times. The
absolute numbers are low for every method, which is the honest summary of the
setting: recovering an exact multi-label set under a source-free domain shift
remains largely unsolved, and the margin between methods is a margin between
partial recoveries.

\begin{figure*}[t]
  \centering
  \small
  \begin{tabular}{@{}l cccc @{}}
  \toprule
  &
  \qualimg{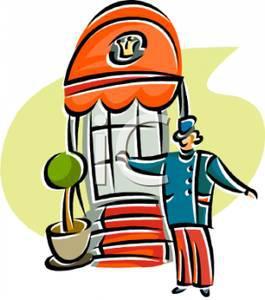} &
  \qualimg{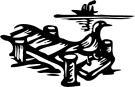} &
  \qualimg{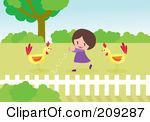} &
  \qualimg{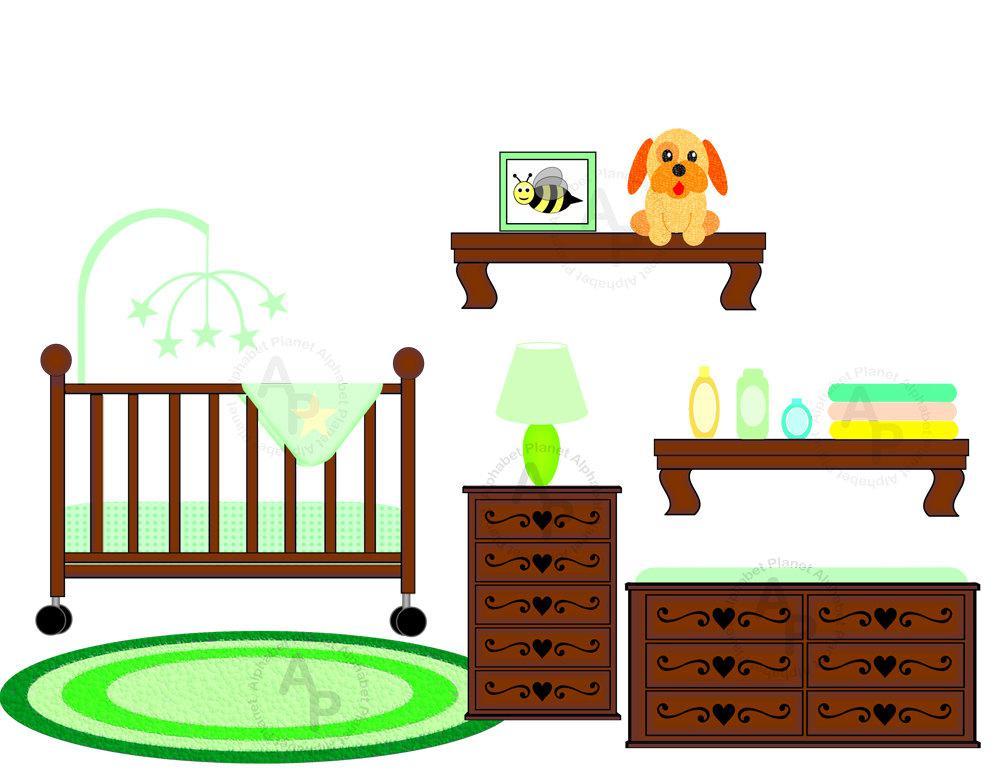} \\[2pt]
  & (a) & (b) & (c) & (d) \\
  \midrule
  Ground truth & \stk{person \\ pottedplant} & \stk{bird \\ boat}
               & \stk{bird \\ person}        & \stk{bottle \\ dog} \\
  \midrule
  Source-only & \stk{chair}  & \stk{chair} & \stk{chair \\ person} & \stk{chair} \\
  SHOT        & \stk{person} & \stk{chair} & \stk{person}          & \stk{bottle \\ chair} \\
  \rev{NOTELA} & \rev{\stk{person}} & \rev{\stk{chair}} & \rev{\stk{person}} & \rev{\stk{bottle \\ chair}} \\
  UCON        & \stk{---}    & \stk{---}   & \stk{person}          & \stk{person} \\
  NVC-LLN     & \stk{---}    & \stk{train} & \stk{---}             & \stk{---} \\
  AC-SFDA     & \stk{person} & \stk{---}   & \stk{person}          & \stk{---} \\
  \ours{} (ours) & \stk{person} & \stk{---}   & \stk{person}          & \stk{bottle} \\
  \bottomrule
  \end{tabular}
  \caption{Four Clipart validation images with the label sets predicted for
  them at threshold $0.5$; ``---'' denotes an empty prediction. The images were
  selected by a rule that references only the annotation: the first four
  carrying exactly two ground-truth labels and pairwise distinct label
  sets.}\label{fig:examples}
\end{figure*}

Figure~\ref{fig:examples} shows what those partial recoveries look like. The examples were chosen by a rule that inspects only the
annotation, never the predictions, and they are accordingly not flattering. On
three of the four images \ours{} recovers one of the two labels and misses the
other; on the fourth it predicts nothing. The failures are shared rather than
particular: no method recovers a complete set on any of the four, \rev{NVC-LLN
is empty on three of them and AC-SFDA on two}, and the unadapted source model
contributes the spurious \emph{chair} that several methods inherit. What separates the methods
here is how often they recover anything at all, which is the same quantity
Table~\ref{tbl:setrecovery} measures over the full split. The panel is included
because it makes the limitation concrete: the gap between a method that helps
and one that does not, in this setting, is a gap between recovering part of a
label set and recovering none of it.

\subsection{Negative result: confidence filtering}\label{sec:negative}

CleanAdapt-TS~\citep{cleanadapt}, whose teacher--student scheme this work follows,
improves its pseudo-labels by discarding the least confident \emph{samples} and
training only on the retained fraction. That mechanism is suited to the
setting it was designed for: when an image has one label, its pseudo-label
is either right or wrong. In the multi-label setting it is blunt, because a
pseudo-label is a vector of $C$ entries of which typically a few are wrong and
the rest are right, so discarding the image discards also some correct entries.

Thus, a label-level variant was implemented, which keeps every image but
masks individual entries: for each class independently, the lowest-scoring
fraction of (sample, class)
pairs is excluded from the loss,
leaving the rest to contribute as usual. Confidence is scored by similarity to
the class prototypes, which is the most effective of the several scoring rules tried.
With the retained fraction ramped from $0.7$ to $0.9$ over the first thirty
epochs, this costs $0.48$ mAP on average relative to no filtering at all
($55.70$ versus $56.18$): $-1.85$ on AID$\rightarrow$UCM, $-0.56$ on
Cityscapes$\rightarrow$Foggy, $-0.41$ on Clipart$\rightarrow$VOC, $-0.01$ on
UCM$\rightarrow$AID, and $+0.45$ on VOC$\rightarrow$Clipart, the single task
where it helps and the one on which the source model transfers worst. These are single
runs, so the per-task figures should be read against the seed spread of $0.4$
to $0.6$ mAP reported in Section~\ref{sec:ablations}; what the experiment
supports is that filtering does not improve on retaining every pseudo-label
\rev{for this scoring rule and this ramp; a different confidence measure or
schedule is not tested here}. All reported results therefore use
$r=1$, which retains every entry.

Two observations make the outcome less surprising. The first is
that the label-level variant is the permissive one: it removes strictly less
supervision than sample-level filtering, so a mechanism that fails here is
unlikely to succeed in its blunter form. The second connects to
Section~\ref{sec:contrastive}. The filter ranks entries by prototype
similarity, and the diagnostic there shows that on tasks where the source
representation already clusters the target data, most of the agreement between
prototype similarity and the labels is agreement the model has generated
itself. Ranking by that quantity therefore selects for consistency with the
model's existing beliefs rather than for correctness, and removes precisely the
entries on which the teacher and the prototypes disagree -- which is where new
information would have to come from. That the one task with a positive result
is also the one where the diagnostic finds prototype similarity most
informative is consistent with this reading, though with a single seed and one
positive case it is a suggestion rather than a finding.

\section{Discussion}\label{sec:discussion}

\paragraph{Why multi-label SFDA is not the union of its parts.}
\rev{The most consequential result of the benchmark is that half of the
extended baselines fail to improve on doing nothing at all. The failures are
not incidental, and the two clearest have separate causes.} SHOT's centroid assignment emits one positive
label per image, so its training signal is single-label regardless. \rev{UCON's
uncertainty control is defined by competition between classes and survives no
per-class substitution, as \ref{app:baselines} sets out.} Neither
mechanism has a multi-label counterpart
that preserves its behaviour\rev{, so in both cases what is measured is the
reach of the single-label formulation rather than a deficiency of the
underlying idea}.

\rev{Two further assumptions break for every method in the comparison rather
than only the baselines: the diversity target of information maximization,
which \ours{} inherits in its miscalibrated form and which helps regardless,
and confidence-weighted prototype construction, which \ours{} replaces. What
all four have in common is only their origin: each is a place where (P1) or
(P2) was doing silent work, and each fails differently once that is withdrawn.
That is why the setting requires the assumption-by-assumption treatment
attempted here rather than a uniform substitution of sigmoids for softmax.}

A second pattern runs through the ablations. The two components that reshape
the target feature space, prototype contrastive alignment and information
maximization, both contribute several times more on VOC$\rightarrow$Clipart
than on Clipart$\rightarrow$VOC, and the prototype initialization follows the
same ordering. The symmetric term, whose role is to bound the influence of
wrong pseudo-labels rather than to reorganize features, does not. The
asymmetry cannot lie in the domain pair, which is identical in both
directions; it lies in where the source model starts, at $39.77$ mAP on
Clipart against $55.52$ on VOC. Removing prototype contrastive alignment on
VOC$\rightarrow$Clipart forfeits $4.70$ of the $5.51$ mAP that adaptation gains
over the source model, against $1.08$ of $7.63$ in the reverse direction. With
two directions this remains just a pattern, but it is the ordering that the
mechanisms claimed in Section~\ref{sec:method} predict, and it \rev{suggests where to
look}: the worse the source model transfers, the larger the share
of the achievable gain that comes from feature-space alignment rather than from
noise robustness. \rev{Source-model mAP is a proxy for how far the
representation has to move and not a measure of the domain gap, since a target
set may simply be harder; the pattern is stated for the two directions
examined and not as a law.}

\paragraph{The gap to multi-label UDA.}
\rev{Section~\ref{sec:benchmark} located the remaining deficit in ranking
quality rather than in where the decision boundary falls. That is a more
specific finding than it first appears, because it raises the possibility that
what is hardest to replace without source data is less the raw supervision
than the label marginal.} A method
with source access observes how often each class occurs and how classes
co-occur, and inherits a usable prior for both. A source-free method has no
comparable signal, which is why the diversity term of
Section~\ref{sec:method} must fall back on a target of $\bar{q}_{c}=0.5$ that
is wrong for nearly every class in these datasets. A
marginal-aware replacement, in which each class's diversity target is an
estimate of its frequency, \rev{is one route to
narrowing the ranking gap, though no variant of it is tested here}. Estimating it from the target data is the difficulty, and prior-correction procedures of the
expectation-maximization family are the natural starting point.

\rev{\paragraph{Reporting the agreement diagnostic elsewhere.}}
\rev{The prototype-agreement diagnostic of
Section~\ref{sec:contrastive} is worth running alongside any prototype-based
adaptation result, not only this one. A final metric can show that a component costs
performance; it cannot show whether the cost arises because the component is
reinforcing the model's own beliefs, and that distinction decides whether the
remedy is to tune the component or to remove it. The measurement is cheap --
two AUCs per refresh over data the method already has in hand -- and needs no
labels at adaptation time, since the ground-truth side is scored afterwards.}

\paragraph{Limitations.}
Several limits of this study should be stated plainly. The ablations are run on
two of the five tasks, and the two VOC$\leftrightarrow$Clipart directions were
chosen because they span the worst and one of the best source-to-target
transfers in the benchmark;
whether the component ordering holds on the remote-sensing and driving tasks is
not established here\rev{, and the diagnostic of
Section~\ref{sec:contrastive} adds only one of them}. All results use a single backbone, so the comparison
isolates adaptation rather than capacity but says nothing about how the
conclusions scale. Two of the design choices examined in
Section~\ref{sec:ablations} are retained for reasons other than
\rev{measured performance}: the
prototypes are kept static as making them learnable is measurably
neutral, and the classifier head is kept frozen although unfreezing it improves
the two-task average by $0.21$ mAP. Both choices are defensible on terms of comparability, but neither is supported by the measurements, and
presenting them otherwise would misrepresent the evidence. \rev{Prototype coverage is incomplete on two tasks: on
VOC$\rightarrow$Clipart four of the twenty classes never exceed $\tau$ during
initialization and so never receive a prototype, as does one of seventeen on
AID$\rightarrow$UCM; the other three tasks cover every class. This is (P2)
acting on the construction itself, and it falls hardest on
VOC$\rightarrow$Clipart, the task where the alignment term contributes most.
\ref{app:tau} shows the cost is nonetheless small: removing the threshold
restores full coverage and loses $1.91$ mAP there, so what the threshold
excludes weighs more than how many anchors it leaves.}
The information
maximization term retains the miscalibrated diversity prior discussed above.
\rev{Prototype contrastive alignment, the largest contributor in
Section~\ref{sec:ablations}, is worth only $+0.86 \pm 0.45$ mAP averaged over the five
tasks (Table~\ref{tbl:contrastive}): it pays on two, is neutral on two and
costs on one. The component ablation is run on the two directions where it
pays, so it overstates what the term contributes benchmark-wide.}
The baselines are multi-label \rev{extensions} constructed for this study rather
than methods published for this setting, so their results characterize how
their mechanisms transfer rather than the performance their authors would
report.

\section{Conclusion}\label{sec:conclusion}

This paper formulated source-free domain adaptation for multi-label
classification, identified the specific assumptions of single-label SFDA that
fail in this setting, and proposed \ours{}, a mean-teacher framework combining
a symmetric per-class objective, label-similarity weighted prototype
contrastive alignment and multi-label information maximization. By adapting
four single-label SFDA methods to multi-label classification \rev{and running
them alongside a published multi-label one}, the first
benchmark for this problem was established across five adaptation tasks and six
datasets. On it, \ours{} achieves the best average mAP and the best \rev{or tied-best} mAP on
\rev{four} of the five tasks, and is the only method to improve on the unadapted source model on
all five; of the four \rev{extended} baselines, two improve on that model on average,
one ends level with it and one collapses far below it. The accompanying analysis shows that prototype contrastive
alignment \rev{contributes unevenly across the five tasks, its largest gain
falling on the transfer where the source model starts weakest}, and
introduces a diagnostic \rev{for separating} genuine alignment from
self-reinforcing agreement -- a distinction that should prove useful beyond
this paper. Future work should target the marginal-aware regularization the
multi-label setting demands, and extend the formulation to open and partial
label sets.

\ifanonymous\else
\printcredits
\fi

\section*{Declaration of generative AI and AI-assisted technologies in the
manuscript preparation process}
During the preparation of this work the authors used Google Gemini 3.1 Pro and
Anthropic Claude Opus 5 in order to refine technical phrasing and support
literature search. After using these tools, the authors reviewed and edited the
content as needed and take full responsibility for the content of the published
article.

\section*{Funding}
This research did not receive any specific grant from funding agencies in the
public, commercial, or not-for-profit sectors.

\section*{Declaration of competing interest}
The authors declare that they have no known competing financial interests or
personal relationships that could have appeared to influence the work reported
in this paper.

\section*{Data availability}
This study uses only publicly available benchmark datasets, which are named and
cited in Section~\ref{sec:setup}; no new data were generated. The
implementation, together with the split files and the evaluation code, is
available at \url{https://github.com/salihnuman/mpact-benchmark}.

\appendix
\renewcommand{\thesection}{Appendix~\Alph{section}}
\counterwithin*{equation}{section}
\counterwithin*{table}{section}
\counterwithin*{figure}{section}
\renewcommand{\theequation}{\Alph{section}.\arabic{equation}}
\renewcommand{\thetable}{\Alph{section}.\arabic{table}}
\renewcommand{\thefigure}{\Alph{section}.\arabic{figure}}
\section{\rev{Multi-label extension of the baselines}}\label{app:baselines}

\rev{This appendix records the per-method changes summarized in
Section~\ref{sec:baselines}, and the reasoning behind the one optimization
setting held common across methods rather than taken from each publication.}

\emph{SHOT}~\citep{liang2020shot} clusters target features around class
centroids and assigns each sample the label of its nearest centroid. The
centroid distance is measured by binary KL divergence in sigmoid space rather than
categorical KL in softmax space, and the information maximization and
pseudo-label losses are kept in their binary form. However, the original $\arg\min$
centroid assignment is retained, so that each image receives exactly one positive label.
This is deliberate: SHOT's pseudo-labeling step has no multi-label formulation
that preserves its published behavior, and reporting the faithful version makes
the structural incompatibility visible rather than hiding it behind a custom
re-design. Pseudo-labels are refined
over three centroid rounds, as per the published configuration.

\emph{UCON}~\citep{ucon} controls uncertainty over memory banks of target
features and predictions. \rev{Its objective has four terms: the two halves of
a contrastive loss, attracting each sample to its $K$ nearest neighbours and
repelling it from the rest of the batch; a dispersion-control term aligning
each sample with an augmented view of itself; and a partial-label loss applied
to uncertain samples.} The score bank stores sigmoid probabilities rather than
softmax distributions; neighbor consistency becomes a binary KL divergence
between a sample's sigmoids and those of its $K$ retrieved neighbors; \rev{the
repulsion term} is computed on $L_{2}$-normalized sigmoids as described above;
and \rev{dispersion control} is a binary cross entropy between the strongly
augmented logits and the detached weakly augmented sigmoids. \rev{The decay
schedule UCON applies to the repulsion coefficient is not reproduced; it is
held at one.} Its uncertainty measure, originally a top-1/top-2
logit ratio, has no meaning once an image may carry several valid labels, so
uncertainty is redefined per class as the distance of the sigmoid output from
the $0.5$ boundary, and ambiguous (sample, class) pairs are masked out of the
pseudo-label loss. \rev{Entries with $|p-0.5|\leq 0.3$ are masked, so the loss
sees only those below $0.2$ or above $0.8$.}

\rev{What UCON does with those ambiguous pairs instead is itself undefined
here, which is why they are excluded rather than relaxed. Its partial-label
loss supervises precisely the uncertain samples, summing a smoothed cross
entropy over a candidate set of their most probable classes. That is a
\emph{relaxation} only because the candidates compete: exactly one is correct,
so spreading mass across them commits to none of them in particular. Under
independent sigmoids they do not compete, and the same sum instead asserts that
every class in the set is present -- a stronger claim than the uncertainty
warranted, not a weaker one. By (P1) the construct loses its meaning rather
than its effectiveness, and no per-class reading recovers it.}

\rev{That substitution is the weakest point of the four extensions. The
original ratio is scale-free: it asks whether the leading
class is clearly ahead of the runner-up, so it keeps discriminating as the
predicted scores shrink. An absolute band cannot, and per class there is no
runner-up to compare against -- which is (P1) in its sharpest form, since the
quantity UCON relies on is defined by competition. The consequence is
measurable. On VOC$\rightarrow$Clipart, $97.8\%$ of (sample, class) entries
fall below $0.2$ and only $2.0\%$ above $0.8$, so $98\%$ of the supervision the
pseudo-label loss receives asserts that a class is \emph{absent}, and the term
reinforces the empty predictions it is meant to correct.}

\rev{Whether that substitution is what holds UCON back was checked directly.
With the partial-label loss removed altogether and everything else unchanged,
UCON averages $30.47 \pm 0.86$ mAP over the five tasks against
$28.17 \pm 1.22$ as reported -- a per-seed difference of $+2.30 \pm 2.05$ --
and stays $21.07$ mAP below the unadapted source model. The criterion
therefore does not survive a per-class reading, and removing the loss it
supervises does not recover the method either: without its central mechanism
UCON does not adapt. Which part of the remaining method the collapse rests on
is not established here.}

\emph{NVC-LLN}~\citep{nvclln} treats adaptation as learning with label noise,
combining contrastive alignment over feature-space neighbors, augmentation
consistency, and a pseudo-label loss down-weighted by sample noisiness. As
above, the score bank holds per-class sigmoids and the dispersion term operates
on $L_{2}$-normalized predictions. For neighbor consistency, binary cross
entropy replaces the original KL divergence, which amounts to summing KL
divergences over $C$ independent Bernoulli distributions. The noise-aware
weighting is re-derived for the multi-label case: each sample's binary entropy
is normalized by its multi-label maximum $C\ln 2$, giving a weight
$1-\mathcal{H}/(C\ln 2)$ that scales that sample's pseudo-label loss. The
implementation deviates from the original in using a single strongly augmented
view rather than two, with the consistency target obtained by
temperature-sharpening the weak view's sigmoids. The neighborhood size is $K=5$.

\emph{AC-SFDA}~\citep{acsfda} is itself a teacher-student method: an offline
model is maintained as an EMA of the online model, and a learned confidence
head on the online model's bottleneck emits a scalar $\alpha_{i}$ per image
that blends the online prediction with a per-image dynamic target stored in a
memory bank, so as to resist memorization of noisy pseudo-labels under the
early-time training phenomenon. The offline model's sigmoids are thresholded at
$0.5$ to form hard multi-label pseudo-labels, the confidence-driven objective
is computed with binary cross entropy, and it is averaged over both batch and
class dimensions rather than summed over classes, so that it stays on the same
scale as the pseudo-label loss. Following the original, adaptation begins with
a ten-epoch standard pseudo-labeling warmup before the interactive phase. The
implementation deviates in one respect: the paper recommends replacing batch
normalization with group normalization at small batch sizes, but batch
normalization is retained here so that every method adapts from identical
BN-pretrained source weights, since substituting group normalization at
adaptation time discards the pretrained normalization statistics.

\paragraph{\rev{Optimization settings.}} \rev{The common learning rate is not
arbitrary.} For UCON and NVC-LLN it is exactly the rate
their publications assign to the classifier and bottleneck, applied to the
backbone as well; only AC-SFDA, which uses Adam at $10^{-6}$ and $10^{-5}$,
sits far from it. Reproducing the published per-group rates was considered and
rejected, because those rates are calibrated to their own setups. Taking them unchanged would scale the backbone
rate down by one to three orders of magnitude while the epoch budget stays
fixed at $45$, leaving the representation close to its source initialization
and measuring how little each method moves rather than how well its objective
transfers. Holding the rate common therefore lets every baseline adapt
substantially. The direction of that choice is not uniformly
favourable and is not claimed to be: a larger step helps a method whose
objective transfers to multi-label data and can equally accelerate the collapse
of one whose objective does not, as Section~\ref{sec:benchmark} finds for UCON.

\rev{What the choice costs the two strongest baselines was measured rather than
argued. Each was re-run on all five tasks at the fixed seed under two
alternatives: Adam at $10^{-5}$, the higher of the two rates AC-SFDA's own
publication uses, and a higher flat SGD rate of $3\times10^{-3}$. NVC-LLN
averages $48.10$ and $53.77$ mAP over the five tasks under the two, against
$54.13$ under the common setting; AC-SFDA averages $51.87$ and $51.16$ against
$52.76$. Both alternatives are worse for both methods, and AC-SFDA loses $0.89$
mAP under the optimizer family its own publication adopts, so the harmonized
rate is not what holds either method back. These are single runs per cell and
vary only the flat rate and the optimizer, not the per-group ratio; they
establish that the common setting is not unfavourable to these two baselines,
not that it is optimal for them.}

\section{\rev{Sensitivity to the prototype confidence threshold}}\label{app:tau}

\rev{The threshold $\tau$ of Section~\ref{sec:protoinit} is argued for rather
than tuned: the sigmoid boundary is where the model stops asserting a class,
which fixes $\tau=0.5$ without reference to target performance.
Table~\ref{tbl:tau} reports what that choice costs. Each arm repeats the two
VOC$\leftrightarrow$Clipart directions over three seeds at the protocol of
Section~\ref{sec:ablations}, varying only $\tau$; the $\tau=0$ arm is the
unthresholded cell of the decomposition in Section~\ref{sec:protoinit}.}

\begin{table*}[t]
\caption{\rev{Effect of the prototype confidence threshold $\tau$, ResNet-101,
three seeds. $\Delta$ is the mean per-seed difference from the default
$\tau=0.5$, as in Figure~\ref{fig:ablation}. \emph{Prototypes} counts the
classes that exceed $\tau$ at least once during initialization and therefore
receive an anchor at all.}}\label{tbl:tau}
\begin{tabular*}{\tblwidth}{@{}LCCCCCC@{}}
\toprule
& \multicolumn{3}{c}{VOC$\rightarrow$Clipart} & \multicolumn{3}{c}{Clipart$\rightarrow$VOC} \\
\cmidrule(lr){2-4}\cmidrule(lr){5-7}
$\tau$ & mAP & $\Delta$ & prototypes & mAP & $\Delta$ & prototypes \\
\midrule
$0.0$ & 43.37\sd{0.67} & $-1.91$\sd{0.29} & $20/20$ & 63.33\sd{0.12} & $+0.18$\sd{0.48} & $20/20$ \\
$0.3$ & 45.15\sd{0.44} & $-0.13$\sd{0.13} & $17/20$ & 63.87\sd{0.03} & $+0.72$\sd{0.63} & $20/20$ \\
$0.5$ & 45.28\sd{0.40} & --               & $16/20$ & 63.15\sd{0.59} & --               & $20/20$ \\
$0.7$ & 45.23\sd{0.32} & $-0.06$\sd{0.45} & $15/20$ & 62.18\sd{0.73} & $-0.97$\sd{0.77} & $18/20$ \\
\bottomrule
\end{tabular*}
\end{table*}

\rev{Having a threshold matters; its value does not. Removing it costs $1.91$
mAP on VOC$\rightarrow$Clipart, while $0.3$, $0.5$ and $0.7$ fall within
$0.13$ of each other there, well inside the seed spread. The choice therefore
sits on a plateau rather than at a tuned optimum, which is what an argument
from the decision boundary rather than from target performance should
produce.}

\rev{Coverage and performance come apart. The unthresholded arm builds every
prototype and is the worst of the four; $\tau=0.7$ builds fifteen of twenty on
VOC$\rightarrow$Clipart and matches the default. What the threshold excludes
therefore matters more than how many anchors it leaves, which is the reading
Section~\ref{sec:protoinit} gives it.}

\rev{No single value is best on both directions: $\tau=0.3$ gains $0.72$ mAP
on Clipart$\rightarrow$VOC and loses $0.13$ on VOC$\rightarrow$Clipart, and
$\tau=0.7$ is neutral on the first and costs $0.97$ on the second. Selecting
between them would require the target labels that (P3) withholds, so the
principled value is kept and the spread recorded here.}

\bibliographystyle{cas-model2-names}
\bibliography{refs}

\end{document}